\documentclass[manuscript, screen]{acmart}
\AtBeginDocument{%
  }

\setcopyright{acmlicensed}
\acmJournal{CSUR}
\acmYear{2025} \acmVolume{1} \acmNumber{1} \acmArticle{1} \acmMonth{1}\acmDOI{10.1145/3744339}

\UseRawInputEncoding
\usepackage{etoolbox}
\usepackage[utf8]{inputenc}

\usepackage{algorithmic}

\usepackage{array}
\usepackage[caption=false,font=normalsize,labelfont=sf,textfont=sf]{subfig}
\usepackage{textcomp}
\usepackage{stfloats}
\usepackage{url}
\usepackage{graphicx}
\usepackage{ragged2e} 

\usepackage{graphicx}  
\definecolor{pinkfootnote}{cmyk}{0,0.97,0.30,0.07}
\usepackage{arydshln}  
\usepackage{multirow}  
\usepackage{amsmath}
\usepackage{siunitx}
\usepackage{tikz}
\usetikzlibrary{positioning, math, fit, backgrounds}
\usepackage{adjustbox}
\usepackage{balance}
\usepackage{svg}
\usepackage{tabularx, soul}
\usepackage{threeparttable}
\usepackage{booktabs} 
\usepackage{caption} 
\usepackage{pifont} 
\usepackage{acronym}
\newacro{iot}[IoT]{internet of things}
\newacro{ai}[AI]{artificial intelligence}
\newacro{mcu}[MCU]{microcontroller}
\newacro{nas}[NAS]{neural architecture search}
\newacro{qat}[QAT]{quantization-aware training}
\newacro{ptq}[PTQ]{post-training quantization}
\newacro{cnn}[CNN]{convolution neural network}
\newacro{dnn}[DNN]{deep neural network}
\newacro{ram}[RAM]{random-access memory}
\newacro{fov}[FOV]{field of view}
\newacro{ste}[STE]{straight-through estimation}
\newacro{bnn}[BNN]{binarized neural network}
\newacro{dl}[DL]{deep learning}
\newacro{od}[OD]{object detection}
\newacro{gpu}[GPU]{graphics processing unit}
\newacro{cpu}[CPU]{central processing unit}
\newacro{yolo}[YOLO]{you only look once}
\newacro{rpn}[RPN]{region proposal network}
\newacro{ssd}[SSD]{single shot multiBox detector}
\newacro{hog}[HOG]{histogram of oriented gradients}
\newacro{dpm}[DPM]{deformable parts model}
\newacro{sift}[SIFT]{scale-invariant feature transform}
\newacro{map}[mAP]{mean average precision}
\newacro{ap}[AP]{average precision}
\newacro{auc}[AUC]{area under curve}
\newacro{iou}[IoU]{intersection over union}
\newacro{flop}[FLOP]{floating point operations per second}
\newacro{sram}[SRAM]{static random access memory}
\newacro{mac}[MAC]{multiply-accumulate operations}
\newacro{ste}[STE]{straight-through estimator}
\newacro{kd}[KD]{knowledge distillation}
\newacro{rl}[RL]{reinforcement learning}
\newacro{ea}[EA]{evolutionary algorithm}
\newacro{enas}[ENAS]{efficient neural architecture search via parameter sharing}
\newacro{snn}[SNN]{spiking neural network}
\newacro{ann}[ANN]{artificial neural network}
\newacro{wsod}[WSOD]{weakly supervised object detection}
\newacro{rnn}[RNN]{recurrent neural network}
\newacro{lgd}[LGD]{label-guided self-distillation}
\newacro{sssd}[SSSD]{smooth and stepwise self-distillation}
\newacro{mse}[MSE]{mean squared error}
\newacro{cadn}[CADN]{category-aware object detection network}
\newacro{fpn}[FPN]{feature pyramid network}
\newacro{nats}[NATS]{neural architecture transformation search}
\newacro{dnas}[DNAS]{differentiable neural architecture search}
\newacro{lstm}[LSTM]{long short-term memory}

\newacro{sgd}[SGD]{stochastic gradient descent}
\newacro{iceds}[ICEDS]{inception convolution with efficient dilation search}
\newacro{edo}[EDO]{efficient dilation optimization}
\newacro{dlb}[DLB]{depthwise linear block}
\newacro{asq}[ASQ]{adaptive scale quantization}
\newacro{sram}[SRAM]{static random-access memory}
\newacro{fomo}[FOMO]{faster objects more objects}
\newacro{ml}[ML]{machine learning}
\newacro{tinyml}[TinyML]{tiny machine learning}

\newacro{rf}[RF]{receptive field}
\newacro{voc}[VOC]{visual object classes}
\newacro{tpu}[TPU]{tensor processing units}
\newacro{svm}[SVM]{support vector machines}
\newacro{vit}[ViT]{vision transformer}
\newacro{r-cnn}[R-CNN]{region-based convolutional neural networks}
\newacro{isa}[ISA]{instruction set architecture}
\newacro{panet}[PANet]{path aggregation network}
\newacro{nn}[NN]{neural network}
\newacro{caq}[CAQ]{classifier adaptive qantization}
\newacro{rvq}[RVQ]{residual vector quantization}
\newacro{admm}[ADMM]{alternative direction method of multipliers}
\newacro{fpga}[FPGA]{field programmable gate array}
\newacro{fqn}[FQN]{fully quantized network}
\newacro{conv}[CONV]{convolutional}
\newacro{fps}[FPS]{frames per second}
\newacro{js}[JS]{jensen shannon}
\newacro{hnas}[HNAS]{hardware aware neural architecture search}
\newacro{hw}[HW]{hardware aware}
\newacro{asic}[ASIC]{application specific integrated circuit}
\newacro{npu}[NPU]{neural processing unit}
\newacro{genius}[GENIUS]{GPT-4 enhanced neural architecture search}
\newacro{llm}[LLM]{large language model}
\newacro{has}[HAS]{hardware aware scaling}
\newacro{coco}[COCO]{common objects in context}
\newacro{voc}[PASCAL VOC]{pascal visual object classes}
\newacro{ilsvrc}[ILSVRC]{imageNet large scale visual recognition challenge}
\newacro{giou}[GIoU]{generalized IoU}
\newacro{diou}[DIoU]{distance IoU}
\newacro{ciou}[CIoU]{complete IoU}
\newacro{tp}[TP]{true positive}
\newacro{fp}[FP]{false positive}
\newacro{fn}[FN]{false negative}
\newacro{vlm}[VLM]{vision-language model}
\newacro{gan}[GAN]{generative adversarial network}
\newacro{reprta}[RepRTA]{re-parameterizable region-text alignment}
\newacro{savpe}[SAVPE]{semantic activated visual prompt encoder}
\newacro{esod}[ESOD]{efficient small object detection}
\newacro{cfu}[CFU]{custom computational function}

\newcommand{\adcomment}[1]{{\color{black}{#1}}}
    
\newcommand{\xmark}{\textcolor{red}{\ding{55}}}
\begin{document}

\title{Designing Object Detection Models for TinyML: Foundations, Comparative Analysis, Challenges, and Emerging Solutions}
\author{Christophe El Zeinaty}
\email{christophe.el-zeinaty@insa-rennes.fr}
\affiliation{%
  \institution{Univ. Rennes, INSA Rennes, CNRS, IETR - UMR 6164}
  \city{Rennes}
  \country{France}
}
\author{Wassim Hamidouche}
\email{wassim.hamidouche@@ku.ac.ae}
\affiliation{%
  \institution{KU 6G Research Center, Khalifa University}
  \city{Abu Dhabi}
  \country{UAE}
}
\author{Glenn Herrou}
\affiliation{%
  \institution{Univ. Rennes, INSA Rennes, CNRS, IETR - UMR 6164}
  \city{Rennes}
  \country{France}
}\
\author{Daniel Menard}
\email{daniel.menard@insa-rennes.fr}
\affiliation{%
  \institution{Univ. Rennes, INSA Rennes, CNRS, IETR - UMR 6164}
  \city{Rennes}
  \country{France}
}\
\renewcommand{\shortauthors}{El Zeinaty et al.}
\begin{abstract}
\Ac{od} has become vital for numerous computer vision applications, but deploying it on resource-constrained \ac{iot} devices presents a significant challenge. These devices, often powered by energy-efficient microcontrollers, struggle to handle the computational load of deep learning-based \ac{od} models. This issue is compounded by the rapid proliferation of \ac{iot} devices, predicted to surpass 150 billion by 2030. TinyML offers a compelling solution by enabling \ac{od} on ultra-low-power devices, paving the way for efficient and real-time \ac{od} at the edge. Although numerous survey papers have been published on this topic, they often overlook the optimization challenges associated with deploying \ac{od} models in TinyML environments.  To address this gap, this survey paper provides a detailed analysis of key optimization techniques for deploying \ac{od} models on resource-constrained devices. These techniques include quantization, pruning, knowledge distillation, and neural architecture search. Furthermore, we explore both theoretical approaches and practical implementations, bridging the gap between academic research and real-world edge \ac{ai} deployment. Finally, we compare the key performance indicators (KPIs) of existing \ac{od} implementations on microcontroller devices, highlighting the achieved maturity level of these solutions in terms of both prediction accuracy and efficiency. \adcomment{We also provide a public repository to continually track developments in this fast-evolving field: }\href{https://github.com/christophezei/Optimizing-Object-Detection-Models-for-TinyML-A-Comprehensive-Survey}{Link}.
\end{abstract}

\ccsdesc[500]{Computing methodologies~Object detection}

\keywords{Model Compression, Embedded Systems, Edge Computing, TinyML}


\maketitle

\section{Introduction}
\begin{figure}[t]
    \centering
    \subfloat[\ac{od} examples showcasing a variety of detected objects and their corresponding bounding boxes.]{%
        \includegraphics[width=0.9\textwidth]{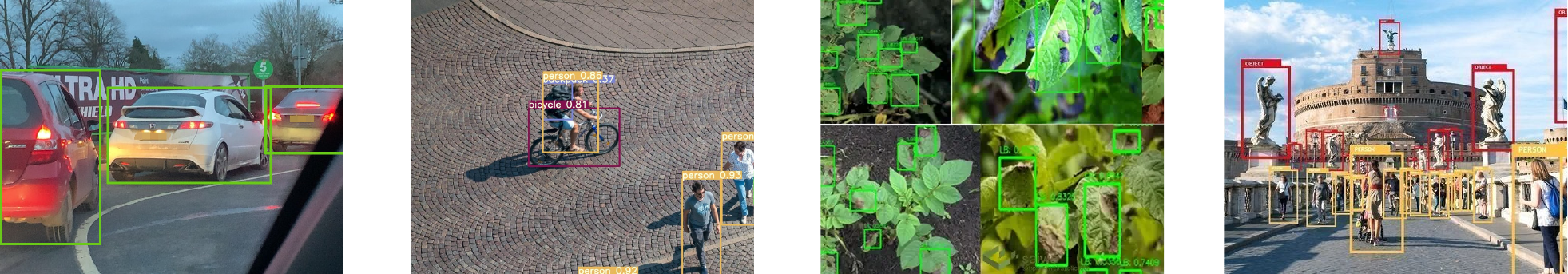}
        \label{fig:pascal-voc}
    } \hfill
    \subfloat[Diverse deployment scenarios of Edge AI and TinyML, including autonomous driving, UAVs, smart farming, and XR glasses illustrating varying compute, memory, and energy constraints across real-world applications.]{%
        \includegraphics[width=0.9\textwidth]{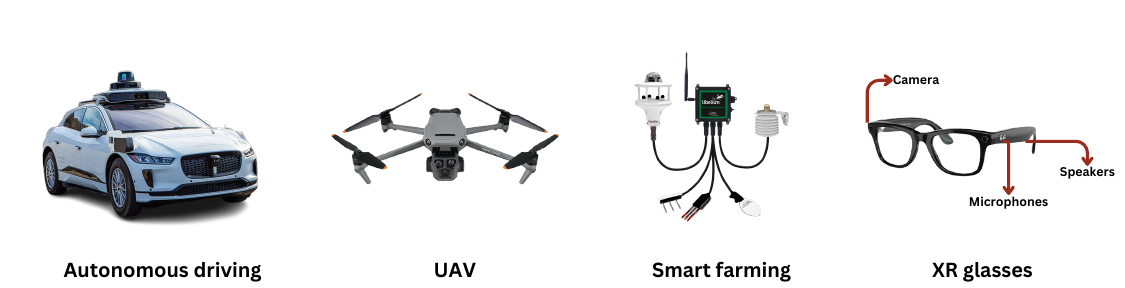}
        \label{fig:usecases}
    }
    \caption{Two examples showcasing different aspects of object detection and their TinyML applications.}
    \label{fig:combined}\vspace{-5mm}
\end{figure}

\begin{figure}[t]
  \centering
  \includegraphics[width=0.95\textwidth]{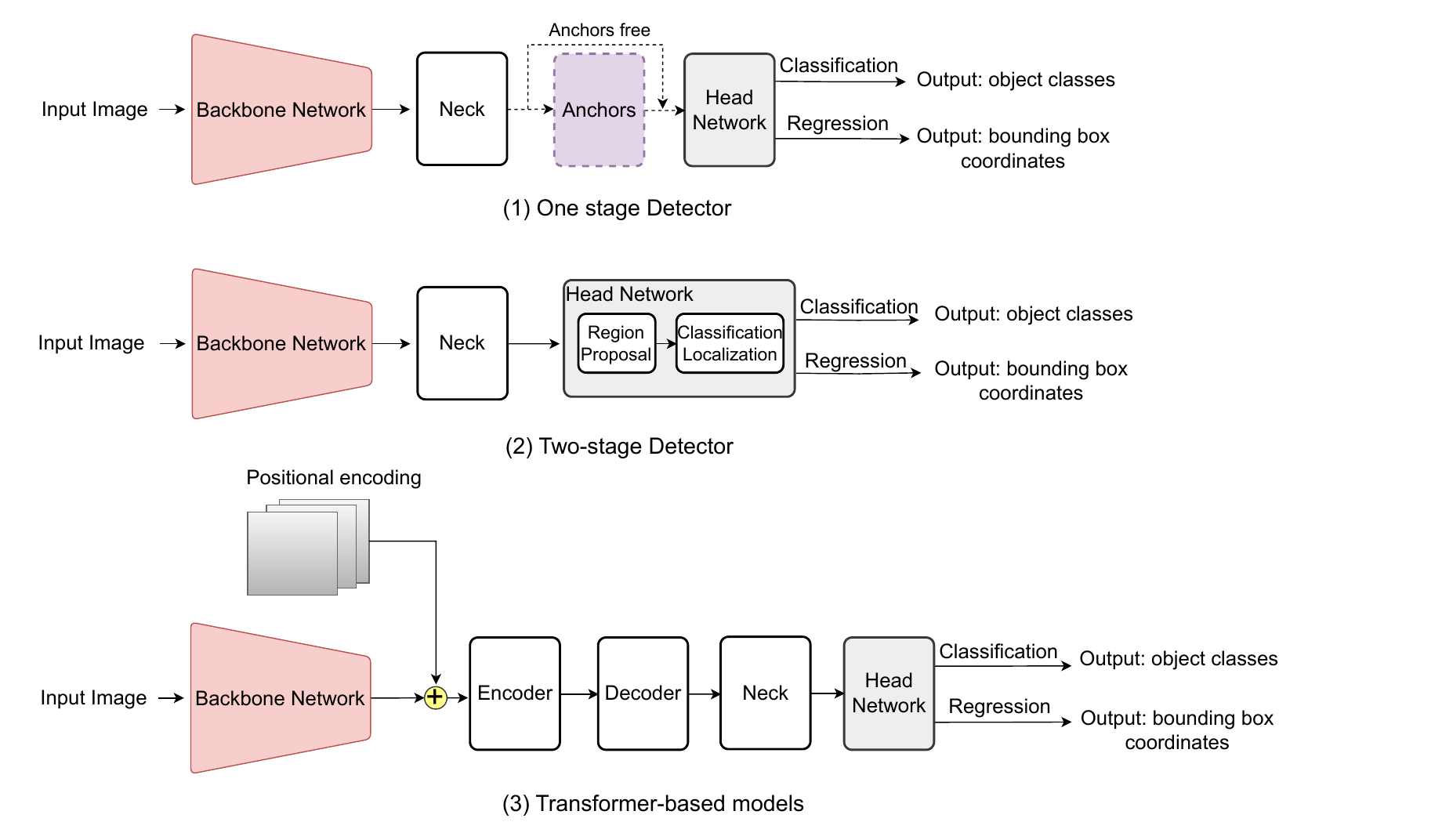}
   \caption{\adcomment{Overview of object detection architectures.}}
  \label{fig:one-two-stage-archi}\vspace{-5mm}
\end{figure}

Technological advancements in computer vision have made \ac{od} a crucial task for applications ranging from autonomous vehicles to medical imaging and security. The primary objectives of \ac{od} are twofold: identifying the location of objects within an image and classifying them according to predefined categories \adcomment{(see Fig.~\ref{fig:pascal-voc})}. Historically, \ac{od} relied on handcrafted features and classical machine learning methods, such as the widely known Viola-Jones detector~\cite{viola-jones} or \ac{hog}~\cite{hog} combined with \ac{svm}. These early approaches were computationally efficient but lacked the accuracy needed for modern, large-scale applications. The advent of \ac{dl} has transformed \ac{od} by enabling the creation of more powerful and accurate models, which typically follow a pipeline consisting of three key components: the backbone, the neck, and the head (see Fig.~\ref{fig:one-two-stage-archi}).  The backbone network is responsible for extracting feature maps from the input image, effectively acting as a feature extractor. Positioned between the backbone and the head, the neck further processes these feature maps.  This often involves techniques like \ac{fpn}~\cite{fpn} or \acp{panet}~\cite{panet}, which  enable the detection of objects at various scales by merging feature maps from different layers of the backbone. Finally, the head is responsible for generating the final predictions, including bounding boxes and object classifications. 
Early \ac{dl}-based methods, such as \ac{r-cnn}~\cite{rcnn}, introduced the concept of region proposal networks to detect objects in multiple stages, incorporating a backbone for feature extraction. This approach was followed by advancements like Faster \ac{r-cnn}~\cite{Faster-RCNN}, which improved efficiency by combining region proposal generation and classification within a single framework. One-stage detectors, such as \ac{yolo}~\cite{yolov1} and \ac{ssd}~\cite{ssd}, further streamlined the pipeline by eliminating the region proposal step, utilizing a backbone, neck, and head to predict both class labels and bounding boxes simultaneously, achieving real-time performance. These models set new standards for \ac{od}, achieving remarkable accuracy and efficiency. However, they are computationally expensive and resource-intensive, making them impractical for deployment on edge devices with limited resources.



The growing proliferation of \ac{iot} devices, forecasted to exceed 150,55 billion such units by 2030~\cite{Kaur2021}, has shifted the focus of \ac{od} research towards more resource-efficient models capable of operating on constrained hardware. Unlike cloud-based solutions, where computational power is abundant, edge devices such as smart cameras, drones, and embedded systems  operate under stringent constraints in terms of memory, processing power, and energy consumption~\cite{Rachmanto, pcamp, Hao-cloud, alqahtani2024benchmarking}. These devices are required to perform \ac{od} tasks in real-time, often with minimal latency, to support applications such as autonomous driving, unmanned aerial vehicle (UAV) based monitoring, smart farming, and extended reality (XR) interaction (see Fig.~\ref{fig:usecases}). As the number of such devices continues to grow exponentially, there is an increasing demand for efficient \ac{od} models that can run locally, thus reducing reliance on cloud computing while addressing issues of latency, bandwidth, and privacy. The emergence of  \ac{tinyml} has opened new avenues for enabling \ac{dl} inference on ultra-low-power devices like \acp{mcu}, which typically operate with less than 1MB of memory. \ac{tinyml} focuses on optimizing \ac{ml} models to fit the constraints of such devices while maintaining acceptable levels of accuracy. Techniques such as quantization, pruning, \ac{kd}, and \ac{nas}~\cite{nagel2021white, NIPS1989_6c9882bb, hinton2015distilling, elsken2019neural} have become central to this effort. \adcomment{Although these optimization techniques are broadly applicable across \ac{tinyml} models, this paper specifically demonstrates their application to \ac{od} models, which serve as an illustrative domain to showcase their effectiveness in real-world, resource-constrained scenarios. This survey is aimed at researchers and practitioners working on optimizing \ac{dl} models for deployment in resource-limited environments, particularly those focused on \ac{od} applications. By structuring our discussion around \ac{od}-specific challenges and optimization strategies, we provide targeted insights that set this work apart from general surveys on lightweight model compression.} These methods enable the deployment of \ac{od} models on highly resource-constrained environments without compromising too much on performance.
While several surveys~\cite{Setyanto, Mittal, Huang, Liu2024, Lin-tiny-survey, Kang, Zou, Syed} have explored various aspects of \ac{od} and edge computing, many lack comprehensive coverage of key optimization techniques and fail to address the specific challenges associated with deploying \ac{od} models in resource-constrained environments, particularly with \ac{tinyml}. Setyanto {\it et al.}~\cite{Setyanto} provide a thorough review of compression techniques across the detection pipeline, focusing on the backbone, head, and neck, but do not delve into the theoretical explanation of broader optimization techniques or \ac{tinyml} deployment. Similarly, Mittal~\cite{Mittal} explores lightweight \ac{od} models and the architecture optimization of the backbone, head, and neck but lacks a detailed analysis of optimization strategies. Huang {\it et al.}~\cite{Huang} offers a more in-depth overview of optimization methods but is limited to the backbone, with little focus on practical deployment scenarios or \ac{tinyml} applications, and without specifics on how these techniques can be applied to \ac{od} in particular. Liu {\it et al.}~\cite{Liu2024} includes a broader discussion of hardware acceleration and compression techniques, touching on \ac{tinyml} challenges, but does not address \ac{od} specific optimization techniques in depth. Lin {\it et al.}~\cite{Lin-tiny-survey} provide comprehensive insights into \ac{tinyml}, including the evaluation of models on \acp{mcu}. However, in their paper, they focused on optimizing a backbone via \ac{nas} and then tested this backbone for an \ac{od} application, making it clear that \ac{od} was not the primary purpose of their work. Additionally, their focus was primarily on their own framework in the explanations provided, rather than offering a broader discussion of optimization techniques. Meanwhile, Kang {\it et al.}~\cite{Kang} provides valuable benchmarking for edge devices but focuses narrowly on the backbone and lacks theoretical explanations of optimization techniques necessary for constrained edge \ac{ai} devices. Similarly, Zaidi {\it et al.}~\cite{Syed} provide a comprehensive review of lightweight networks for edge devices but only discuss backbone architectures and do not explore optimization techniques. Zou {\it et al.}~\cite{Zou} offer another review of edge \ac{ai} with a focus on backbones, but they also lack a discussion on optimization techniques for \ac{od} in \ac{tinyml} environments.

\adcomment{In this paper, we present a survey that extends beyond previous works by covering the full range of optimization techniques for \ac{od} models, with \ac{od} serving as a representative domain within \ac{tinyml} environments.} A detailed examination of novel optimization strategies, tailored to the unique constraints of \acp{mcu} with limited power and memory, is provided, bridging the gap between theoretical understanding and practical deployment in edge \ac{ai} applications~\cite{celzeinaty}. As summarized in Table~\ref{table:survey}, this comparative analysis highlights the key focus areas of prior surveys, illustrating the gaps that our work aims to fill.
\begin{table}[t]
\centering
\resizebox{0.85\columnwidth}{!}{
\begin{threeparttable}
\captionsetup{justification=centering} 
\caption{Recent surveys on model compression techniques and their applications in embedded systems.}
\begin{tabular}{l l c c c c c c}
\toprule
\textbf{Authors} & \textbf{Year} & \textbf{Focus on \ac{od}} & \textbf{Optim. parts} & \textbf{Optim. insights\tnote{$\dagger$}} & \textbf{Hardware Discussion} & \textbf{Edge \ac{ai}} & \textbf{\ac{tinyml}} \\ 
\midrule
Zaidi {\it et al.}~\cite{Syed}          & 2021 & \ \adcomment{\checkmark} & Backbone                 & \xmark & \xmark & \ \adcomment{\checkmark} & \xmark \\ 
Kang {\it et al.}~\cite{Kang}           & 2022 & \ \adcomment{\checkmark} & Backbone                 & \xmark & \ \adcomment{\checkmark} & \ \adcomment{\checkmark} & \xmark \\ 
Zou {\it et al.}~\cite{Zou}             & 2023 & \ \adcomment{\checkmark} & Backbone                 & \xmark & \xmark & \ \adcomment{\checkmark} & \xmark \\ 
Setyanto {\it et al.}~\cite{Setyanto}   & 2024 & \ \adcomment{\checkmark} & Backbone, head, and neck & \xmark & \ \adcomment{\checkmark} & \ \adcomment{\checkmark} & \xmark \\ 
Mittal~\cite{Mittal}                    & 2024 & \ \adcomment{\checkmark} & Backbone, head, and neck & \xmark & \ \adcomment{\checkmark} & \ \adcomment{\checkmark} & \xmark \\ 
Huang {\it et al.}~\cite{Huang}         & 2024 & \xmark    & Backbone                 & \ \adcomment{\checkmark} & \ \adcomment{\checkmark} & \ \adcomment{\checkmark} & \xmark \\ 
Liu {\it et al.}~\cite{Liu2024}         & 2024 & \xmark    & Backbone                 & \ \adcomment{\checkmark} & \ \adcomment{\checkmark} & \ \adcomment{\checkmark} & \ \adcomment{\checkmark} \\ 
Lin {\it et al.}~\cite{Lin-tiny-survey} & 2024 & \xmark    & Backbone                 & \xmark & \ \adcomment{\checkmark} & \ \adcomment{\checkmark} & \ \adcomment{\checkmark} \\ 
\textbf{Ours}                           & 2025 & \ \adcomment{\checkmark} & Backbone, head, and neck & \ \adcomment{\checkmark} & \ \adcomment{\checkmark} & \ \adcomment{\checkmark} & \ \adcomment{\checkmark} \\ 
\bottomrule
\end{tabular}
\label{table:survey}
\begin{tablenotes}
\item[$\dagger$] The \textbf{Opt. Insights} column indicates whether the survey discusses in-depth optimization techniques such as quantization, pruning, \ac{kd}, and \ac{nas} theoretically.
\end{tablenotes}\vspace{-5mm}
\end{threeparttable}
}
\end{table}
The remainder of this paper is structured as follows. In Section~\ref{sec:object-detection-ov}, an introduction to traditional and \ac{dl}-based \ac{od} methods is presented, outlining the transition from handcrafted features to modern \acp{dnn}. Various datasets are discussed in Section~\ref{sec:object-detection-data-sets}, while the critical evaluation metrics commonly used in \ac{od} research are detailed in Section~\ref{sec:object-detection-data-eval} followed by a compartive analysis of different \ac{tinyml} hardware in Section~\ref{sec:tinyml-hardware}. An in-depth discussion of optimization strategies for \ac{od} models in resource-constrained environments is provided in Section~\ref{sec:object-detection-optim}. Furthermore, the deployment of \ac{od} models on embedded systems, with a particular focus on \ac{tinyml}, is explored in Section~\ref{sec:object-detection-embed-app}. Finally, the current challenges and future research directions in this area are highlighted in Section~\ref{sec:object-detection-challenges}, followed by the conclusion in Section~\ref{sec:conclusion}.
\section{Introduction to Object Detection}
\label{sec:object-detection-ov}
\adcomment{\Ac{od} has undergone significant evolution, transitioning from early handcrafted feature-based methods to modern \ac{dl}-driven approaches. While comprehensive surveys have explored advancements in \ac{od} in depth, this section aims to provide a structured introduction that contextualizes the optimization strategies discussed later in the paper. By outlining key developments, we emphasize how \ac{od} techniques have been shaped by the need for computational efficiency, particularly in resource-constrained environments such as embedded systems and \ac{tinyml} applications.} 

In the realm of computer vision, \ac{od} stands as a pivotal task, representing the fundamental capability to identify and precisely locate objects within digital images or video frames. Unlike mere image classification, which assigns a single label to an entire image or a patch, \ac{od} ventures further. It not only discerns the categories of objects present but also draws bounding boxes around each object, pinpointing their spatial coordinates within the visual field. This dual functionality empowers computer systems to not only recognize objects but to understand their spatial context, opening the door to a multitude of applications that require nuanced scene understanding, making \ac{od} essential for applications in autonomous driving, surveillance, medical imaging, and robotics. \adcomment{In \ac{tinyml}, a common application is person detection, which simplifies the task by focusing on binary classification: determining whether a person is present or not, without requiring localization. This makes person detection less computationally intensive than full \ac{od}, which involves both classification and localization. While person detection is widely used in real-time applications such as surveillance, where precise localization may not always be necessary, the ability of \ac{od} to accurately localize objects remains critical in more complex industrial use cases. Benchmarks like MLPerf Tiny~\cite{banbury2021mlperf}, which include person detection as one of the use cases, underscore the importance of efficient models for resource-constrained devices like \acp{mcu}. This highlights the value of studying and optimizing \ac{od} on such devices, where both real-time performance and localization accuracy are essential.}

The inception of \ac{od} can be attributed to the pioneering work of Viola and Jones, who introduced the groundbreaking Viola-Jones framework~\cite{viola-jones}, a significant milestone in computer vision. This framework employed Haar-like features~\cite{haar-like} and a cascade classifier to achieve real-time face detection~\cite{cascade-face}. Haar-like features are simple rectangular filters that efficiently detect variations in brightness, while the cascade classifier enabled rapid rejection of non-object regions, leading to substantial speed improvements. Despite its breakthrough in real-time face detection, early \ac{od} methods, including the Viola-Jones framework, faced challenges in handling complex object categories and variations in object appearance. These limitations spurred the quest for more sophisticated approaches as computer vision tasks grew in complexity.

The landscape of \ac{od} underwent a significant transformation with the emergence of \acp{dnn}. \Ac{dl} techniques showcased remarkable capabilities in feature learning and abstraction, rendering them highly suitable for \ac{od} tasks. \Ac{dnn}, particularly \acp{cnn}, brought about a paradigm shift in \ac{od}. They not only surpassed traditional methods in performance but also enabled end-to-end learning, eliminating the need for handcrafted features. \Ac{cnn}-based \ac{od} models became adept at automatically learning discriminative features from data, resulting in substantial improvements in detection accuracy.

In recent years, the architecture of \ac{od} models has evolved significantly, with major developments categorized into traditional methods and \ac{dl}-based methods.

\subsection{Traditional Methods}
Traditional \ac{od} methods, such as the Viola-Jones framework, \ac{hog}~\cite{hog}, and \ac{dpm}~\cite{dpm}, relied on handcrafted features and simple classifiers. Nevertheless, these methods often struggled with the complexity and variability of real-world objects.

In the 1990s, early \ac{od} algorithms were built upon handcrafted features due to the absence of effective learned image representation techniques at that time. These algorithms necessitate sophisticated feature engineering and various acceleration techniques to achieve acceptable performance. In 2001, Viola and Jones~\cite{viola-jones} achieved real-time detection of human faces using a 700-MHz Pentium III CPU, making it tens or even hundreds of times faster than previous methods. Their approach used a sliding window technique to scan all possible locations and scales in an image for human faces. They dramatically improved detection speed by incorporating three key techniques: the "integral image," "feature selection," and "detection cascades."

In 2005, Dalal and Triggs~\cite{hog} introduced the \ac{hog} descriptor, which improved upon previous feature descriptors like the \ac{sift} and shape contexts. \Ac{hog} balanced feature invariance and nonlinearity, primarily used for pedestrian detection. The detector rescales the input image multiple times to detect objects of different sizes while keeping the detection window size unchanged. The \ac{hog} detector became foundational for many object detectors and computer vision applications. In 2008, Felzenszwalb {\it et al.}~\cite{dpm} proposed the \ac{dpm}, which extended the \ac{hog} detector using a "divide and conquer" approach. This method involves decomposing objects into parts (e.g., a car's window, body, and wheels) and detecting these parts individually. Girshick ~\cite{girshick2011} subsequently extended the \ac{dpm} framework to incorporate "mixture models," enhancing its capacity to handle the greater variability inherent in real-world objects. The \ac{dpm}'s influence reverberated through the development of many modern detection techniques, contributing valuable insights such as mixture models~\cite{Viroli}, hard negative mining~\cite{SouYoung}, bounding box regression~\cite{Ma}, and context priming~\cite{torralba_contextual_2003}. In recognition of their seminal work, Felzenszwalb and Girshick were honored with the "lifetime achievement" award from \ac{voc} in 2010.

\subsection{Deep Learning-based Methods}
\ac{dl}-based methods, utilizing \acp{dnn} such as \acp{cnn}, have revolutionized \ac{od} by facilitating end-to-end learning and feature extraction. This paradigm shift has led to substantial improvements in the accuracy and efficiency of \ac{od} systems, empowering them to learn intricate features directly from the data. Within the realm of \ac{dl}, \ac{od} models can be broadly categorized into two architectural designs: two-stage detectors and one-stage detectors, as illustrated in Fig.~\ref{fig:one-two-stage-archi}.

\subsubsection{Two-Stage Detectors}
Two-stage detectors, exemplified by models such as \ac{r-cnn}~\cite{rcnn} and Faster \ac{r-cnn}~\cite{Faster-RCNN}, employ a two-step process. In the initial stage, these models generate a set of region proposals that are likely to contain objects of interest. Subsequently, in the second stage, these proposals undergo classification and refinement to achieve precise object localization. This approach allows higher detection accuracy, as the model can concentrate its computational resources on promising image regions. However, the additional processing step often leads to slower inference times compared to one-stage detectors.
\subsubsection{One-Stage Detectors}
One-stage detectors, represented by models such as \ac{yolo}~\cite{yolov1}, and \ac{ssd}~\cite{ssd}, directly predict object categories and bounding boxes in a single evaluation step. These models are optimized for real-time applications due to their high speed and efficiency, eliminating the need for a separate region proposal stage. However, while one-stage detectors generally offer faster inference, they may exhibit a slight trade-off in accuracy compared to their two-stage counterparts.

Earlier one-stage methods primarily relied on anchor boxes (i.e., prior boxes) to provide reference points for classification and regression  tasks. However, the inherent variability of objects in terms of number, location, scale, and aspect ratio often necessitated the use of a large number of reference boxes to achieve a close match with ground-truth annotations and attain high performance. To address this limitation, several anchor-free detectors, such as~\cite{centernet}, and CornerNet~\cite{cornernet}, were proposed. These models eliminate the reliance on predefined anchor boxes, thereby simplifying the detection pipeline and reducing computational overhead. By design, anchor-free models offer improved generalization and flexibility in detecting objects of varying sizes and shapes.

Recently, transformer-based models such as DETR~\cite{detr} and RT-DETR~\cite{rtdetr} have emerged, leveraging attention mechanisms to improve detection performance and robustness. These models directly model spatial relationships between different image regions, enabling a more sophisticated contextual understanding of \ac{od}. \adcomment{Expanding on this paradigm, Grounding DINO 1.5~\cite{ren2024grounding} introduces linguistic grounding into \ac{od}, significantly enhancing flexibility in open-set scenarios. The Grounding DINO 1.5 Pro model scales both the architecture and training datasets to over 20 million images, achieving state-of-the-art zero-shot transfer capabilities. Meanwhile, the Edge variant, optimized for real-time applications, achieves 75.2 \ac{fps} with an input size of 640$\times$640 using a TensorRT-optimized pipeline without compromising detection robustness. These advancements underscore the shift towards more adaptable and efficient models for both closed-set and open-world \ac{od} applications.}  Building on the query-based detection framework established by DETR~\cite{detr}, diffusion models~\cite{NEURIPS2019_3001ef25} have introduced a novel approach to \ac{od}, further streamlining the detection pipeline. Wang {\it et al.}~\cite{wang2022diffusiondet} proposed DiffusionDet, which formulates detection as a denoising diffusion process, iteratively refining random initial boxes into precise object locations. Unlike traditional models that depend on predefined object priors, DiffusionDet eliminates hand-designed queries and learnable object representations, offering a simplified yet highly effective detection pipeline that pushes the boundaries of \ac{od} performance. \adcomment{Beyond traditional \ac{od}, generative AI is increasingly being explored for zero-shot detection tasks. RONIN~\cite{nguyen2024zero}, for instance, introduces a diffusion-based zero-shot object-level out-of-distribution (OOD) detection framework. By leveraging context-aware inpainting, RONIN detects whether an object is in or out-of-distribution without requiring access to the original training set, making it particularly suited for black-box models where training data is unavailable. This generative approach further exemplifies how diffusion models can enhance object-level understanding in a zero-shot setting. Meanwhile, the YOLO family continues to evolve, with recent advancements such as YOLOv12~\cite{tian2025yolov12} and YOLO-E~\cite{wang2025yoloe} enhancing real-time \ac{od}. YOLOv12 improves detection efficiency through dynamic label assignment and attention-based mechanisms, while YOLOE extends the traditional YOLO framework with open-set prompt mechanisms, including text and visual prompts, broadening its applicability beyond predefined categories. Additionally, YOLOE integrates novel components like \ac{reprta} and \ac{savpe}, which strengthen vision-language alignment and enable robust zero-shot detection. These advancements reinforce the trend toward highly adaptable and efficient models capable of operating in both constrained and open-world settings. Further bridging the gap between convolutional and transformer-based architectures, hybrid models like RT-DETR~\cite{rtdetr} and YOLO-World~\cite{cheng2024yolo} combine the efficiency of convolutional backbones with the contextual learning power of transformers. This fusion balances speed, accuracy, and generalization, making them promising candidates for diverse \ac{od} applications.} By leveraging advances in transformers, diffusion models, convolutional networks, and hybrid architectures, modern \ac{od} systems continue to enhance efficiency and adaptability, paving the way for further breakthroughs in real-world detection tasks.

\subsection{Loss Optimisation}
\ac{od} models generally strive to optimize a combination of classification and localization tasks. The classification task entails predicting the category of objects present within proposed regions, whereas the localization task involves predicting the precise coordinates of the corresponding bounding boxes. Recent models leverage \ac{dl} to learn these tasks jointly, thereby achieving substantial improvements in accuracy and efficiency compared to traditional methods.

Mathematically, the overall loss function \( L \) employed for training an \ac{od} model is typically formulated as a weighted sum of two components: the classification loss \( L_{cls} \) and the localization loss \( L_{loc} \). \adcomment{To allow for a more flexible balance between these two components, we introduce a hyperparameter \( \beta \) for the classification loss and \( \alpha \) for the localization loss. This relationship is expressed as follows:}  
\begin{equation}  
L = \adcomment{\beta \, L_{cls}} + \alpha L_{loc},  
\end{equation}  
\adcomment{where \( \alpha \) and \( \beta \) are hyperparameters that determine the relative importance of the localization and classification loss components, respectively.}
\subsubsection{Classification Loss}
The classification loss \( L_{cls} \) is typically computed using the cross-entropy loss function, which quantifies the discrepancy between the predicted class probabilities and the corresponding ground-truth class probabilities. For each predicted bounding box, the classification loss can be formally defined as:
\begin{equation}
L_{cls} = -\sum_{i=1}^{C} y_{i} \log(\hat{p}_{i})
\end{equation}
where \( C \) is the total number of classes, \( y_{i} \) represents the ground-truth probability for class \( i \) (1 for the correct class and 0 for others), and \( \hat{p}_{i} \) denotes the predicted probability for class \( i \).

\adcomment{In addition to the standard cross-entropy loss, several variations are commonly used to address issues such as class imbalance and model overfitting. One such variation is Focal Loss, which down-weights the easy examples and focuses more on hard-to-classify examples. This is particularly useful in \ac{od} tasks where certain classes may dominate the training data, causing the model to overlook harder, less frequent classes. Focal loss can be expressed as:}

\begin{equation}
\adcomment{L_{\text{focal}} = -\sum_{i=1}^{C} \alpha_{i} (1 - \hat{p}_{i})^\gamma y_{i} \log(\hat{p}_{i})}
\end{equation}

\adcomment{where \( \alpha_{i} \) is a balancing factor for class \( i \), and \( \gamma \) is a focusing parameter that reduces the relative loss for well-classified examples. Another common approach is Label Smoothing, which modifies the ground-truth labels by softening the targets. This can help prevent the model from becoming too confident about its predictions and improve generalization. Label smoothing can be incorporated into the cross-entropy loss by adjusting the ground-truth labels \( y_i \) to }

\begin{equation}
\adcomment{\tilde{y}_i = (1 - \epsilon) y_i + \frac{\epsilon}{C}},
\end{equation}

\adcomment{where \( \epsilon \) is the smoothing factor and \( C \) is the number of classes. These modifications to the classification loss function allow the model to perform better on imbalanced datasets, reduce overfitting, and improve generalization to unseen examples.}

\subsubsection{Localization Loss}
The localization loss \( L_{loc} \) measures the error in the predicted bounding box coordinates. A common choice is the Smooth L1 loss (Huber loss), which is less sensitive to outliers than the L2 loss. For a predicted bounding box with coordinates \( (\hat{x}, \hat{y}, \hat{w}, \hat{h}) \) and the ground-truth box \( (x, y, w, h) \), the Smooth L1 loss is defined as follows:
\begin{equation}
L_{loc} = \sum_{i \in \{x, y, w, h\}} S(\hat{i} - i),
\end{equation}

where \( S(x) \) is the Smooth L1 function, defined as:
\begin{equation}
S(x) = \begin{cases} 
0.5x^2 & \text{if } |x| < 1, \\
|x| - 0.5 & \text{otherwise}.
\end{cases}
\end{equation}
\adcomment{In addition to Smooth L1 loss, several other regression loss functions are commonly used in \ac{od} to improve bounding box prediction accuracy. For instance, L2 loss (Euclidean distance) is a basic method but can be more sensitive to outliers. It computes the Euclidean distance between the predicted and ground-truth bounding box coordinates. While simple and effective in many cases, it may be more sensitive to large errors than Smooth L1 loss, especially when the predicted bounding boxes are far from the ground truth. Furthermore, IoU-based loss functions, such as \ac{giou}, \ac{diou}, and \ac{ciou}, have been proposed to directly optimize the overlap between predicted and ground-truth boxes, thereby improving localization performance. These functions focus on increasing the overlap between the predicted and ground-truth boxes. For example, GIoU extends IoU by penalizing the distance between the bounding boxes centers and increasing robustness to boxes with no overlap. Similarly, DIoU and CIoU refine bounding box predictions by incorporating additional geometric information, such as aspect ratio and center distance.
}

In summary, \ac{od} models employ a multi-task loss function that effectively balances the objectives of classification accuracy and localization precision. This dual optimization empowers the models to not only identify the categories of objects present in an image but also to accurately pinpoint their spatial locations within the visual scene.
\begin{table}[t]
\centering
\caption{Comparison of Object Detection Datasets}
\resizebox{0.75\columnwidth}{!}{
\begin{tabular}{l c c c c c c c}
\hline
\textbf{Dataset} & \textbf{Year} & \textbf{Classes} & \multicolumn{2}{c}{\textbf{Train}} & \multicolumn{2}{c}{\textbf{Validation}} & \textbf{Test} \\
\cline{4-7}
 & & & \textbf{Images} & \textbf{Objects} & \textbf{Images} & \textbf{Objects}\\
\hline
PASCAL VOC~\cite{pascal-voc-2007} & 2012 & 20 & 5,717 & 13,609 & 5,823 & 13,841 & 10,991\\
ILSVRC~\cite{ilsvrc} & 2015 & 200 & 456,567 & 478,807 & 20,121 & 55,501 & 40,152\\
MS-COCO~\cite{coco-dataset} & 2017 & 80 & 118,287 & 860,001 & 5,000 & 36,781 & 40,670\\
OpenImage~\cite{openimage} & 2020 & 600 & 1,743,042 & 14,610,229 & 41,620 & 204,621 & 125,436\\
\hline  \vspace{-5mm}
\end{tabular}
\label{od-datasets}}
\end{table}

\section{Datasets}
\label{sec:object-detection-data-sets}
\subsection{\adcomment{Object Detection}}
For \ac{od}, diverse and well-annotated datasets are essential as they provide the foundation for training, evaluating, and benchmarking models. Table~\ref{od-datasets} summarizes several well-known datasets frequently used in \ac{od} research. In this review, the primary focus will be on two prominent datasets that have significantly contributed to the advancement of the field:

\noindent \textbf{\Ac{coco}~\cite{coco-dataset}.} \ac{coco} is a widely recognized dataset in the computer vision community. It contains a vast collection of images, each meticulously annotated with object categories and precise object bounding box coordinates. \ac{coco}'s richness lies not only in the diversity of objects but also in the complexity of scenes, making it an invaluable resource for training and evaluating \ac{od} models. With 80 object categories and over 330,000 images, including more than 1.5 million annotated instances, \ac{coco} has become a benchmark for measuring the performance of modern \ac{od} algorithms. 

\noindent 
\textbf{\Ac{voc}~\cite{pascal-voc-2007}.} The \ac{voc} dataset stands as another seminal benchmark that has played a pivotal role in the advancement of \ac{od} research. Featuring a diverse range of object categories, it provides annotated images accompanied by corresponding object bounding boxes. \ac{voc} has served as a standard benchmark dataset for numerous years, enabling the rigorous evaluation and comparison of various \ac{od} methods. The dataset encompasses 20 object categories, with approximately 11,530 images containing 27,450 annotated objects. While it may not be as extensive as the \ac{coco} dataset, its historical significance and contribution to shaping the early years of \ac{od} research remain undeniable. In addition to these two datasets, large-scale datasets such as {\Ac{ilsvrc}}~\cite{ilsvrc} and {OpenImage}~\cite{openimage} have also contributed significantly to \ac{od} research. \ac{ilsvrc}, with 200 object categories and over 1.2 million training images, has been instrumental in advancing both classification and detection tasks. Similarly, the OpenImage dataset, with its vast 600 object categories and more than 9 million annotated objects, is frequently used for training models that generalize well to real-world \ac{od} scenarios. Although this survey focuses on \ac{coco} and \ac{voc}, the availability of large-scale datasets like \ac{ilsvrc} and OpenImage has broadened the scope of \ac{od} research, facilitating improvements in model robustness and scalability.

\subsection{\adcomment{Other \ac{tinyml} Datasets}}
\adcomment{While this paper primarily focuses on \ac{od} models for object classification and localization within images, it is essential to highlight other well-established use cases in \ac{tinyml}. As discussed in~\cite{banbury2021mlperf}, some of the most prominent applications include person detection, image classification, keyword spotting, and anomaly detection. Table~\ref{tinyml-datasets} provides a summary of these key use cases along with the most widely used datasets in each domain.}

\noindent \adcomment{\textbf{Speech Commands Dataset~\cite{speechcommands}.} The Speech Commands dataset is designed for keyword spotting, enabling efficient on-device speech recognition. Released by Google Brain in 2018, it contains 105,829 utterances from 2,618 speakers, covering 35 words. The dataset primarily supports small-footprint models for detecting predefined words such as "Yes," "No," "Up," "Down," and digits. It also includes background noise samples to improve robustness against false activations. The dataset has been widely adopted in \ac{tinyml} applications, particularly for wake word detection in smart assistants.}
\begin{table}[t]
\centering
\caption{Comparison of TinyML Datasets}
\resizebox{0.95\columnwidth}{!}{
\begin{tabular}{l c c c c}
\toprule
\textbf{Dataset Name} & \textbf{Year} & \textbf{Use Case} & \textbf{Number of Samples} & \textbf{Evaluation Metric} \\
\midrule
Speech Commands~\cite{speechcommands} & 2018 & Keyword Spotting & 105,829 & Accuracy \\
ToyAdmos Dataset~\cite{toyadmos} & 2019 & Sound Anomaly Detection & 5,939 & AUC-ROC, Precision-Recall \\
VWW Dataset~\cite{vww} & 2019 & Person Detection & 115,000 & Accuracy \\
Wake Vision Dataset (Quality)~\cite{Njor} & 2024 & Person Detection & 1,322,574 & Accuracy \\
Wake Vision Dataset (Large)~\cite{Njor} & 2024 & Person Detection & 5,760,428 & Accuracy \\
\bottomrule
\end{tabular}
}
\label{tinyml-datasets}
\end{table}

\noindent \adcomment {\textbf{Visual Wake Words (VWW) Dataset~\cite{vww}.} The Visual Wake Words dataset is specifically designed for \ac{tinyml} applications, particularly for low-power vision models deployed on \acp{mcu}. Introduced by Google Research in 2019, the dataset addresses the common \ac{tinyml} use case of determining whether a person is present in an image or not. It is derived from the \ac{coco} dataset by filtering and relabeling images based on the presence of a person occupying at least 0.5\% of the image area. The dataset contains approximately 115,000 images split into training and validation sets, making it a realistic benchmark for evaluating tiny vision models with extreme memory constraints (e.g., models under 250 KB). VWW has been widely adopted in \ac{tinyml} benchmarks to assess the trade-offs between model accuracy, memory footprint, and computational efficiency on \acp{mcu}.}

\noindent \adcomment{\textbf{Wake Vision Dataset~\cite{Njor}.} Wake Vision is a large-scale dataset specifically designed for person detection in \ac{tinyml} applications. It contains over 5 million images and is provided in two variants: Wake Vision (Large) and Wake Vision (Quality). The large variant focuses on dataset scale for pretraining and \ac{kd}, while the quality variant prioritizes high-quality labels for final model performance. The dataset is derived from Open Images v7 and improves accuracy by 1.93\% over existing datasets such as Visual Wake Words (VWW). Additionally, Wake Vision introduces a benchmark suite evaluating model robustness across real-world conditions, including lighting variations, camera distances, and demographic characteristics. 
}

\noindent \adcomment{\textbf{ToyADMOS Dataset~\cite{toyadmos}.} The ToyADMOS dataset is a large-scale dataset designed for anomaly detection in machine operating sounds (ADMOS), making it highly relevant for \ac{tinyml} applications. Released in 2019, it includes over 180 hours of normal machine-operating sounds and more than 4,000 samples of anomalous sounds. The dataset is divided into three sub-datasets: (1) product inspection using a toy car, (2) fault diagnosis of a fixed machine using a toy conveyor, and (3) fault diagnosis of a moving machine using a toy train. Sounds were recorded with four microphones at a 48-kHz sampling rate under controlled conditions. Anomalous sounds were generated by deliberately damaging machine components, allowing for realistic fault detection evaluation. The dataset is also used in the MLPerf Tiny benchmark suite to evaluate \ac{tinyml} models for sound anomaly detection, demonstrating its importance in resource-constrained machine learning applications.}

\section{Evaluating Object Detection Models}
\label{sec:object-detection-data-eval}
\adcomment{In this section, the evaluation criteria and metrics employed to assess the performance of \ac{od} models are examined, highlighting the compromise between quality and cost. The metrics used to assess general performance in terms of \ac{od} accuracy are first presented (\S~\ref{subsec:accu}), followed by an exploration of the efficiency metrics tailored for embedded devices, where both aspects will be discussed (\S~\ref{subsec:effi}).}

\subsection{Accuracy Performance}
\label{subsec:accu}
In the context of \ac{od}, \ac{ap} and \ac{map} are commonly employed in conjunction to evaluate the performance of models. \Ac{ap} is calculated by plotting the precision-recall curve for each individual class. This curve graphically depicts the trade-off between precision (the proportion of true positive detections among all positive detections) and recall (the proportion of true positive detections among all actual positive instances) at varying confidence thresholds. In \ac{od}, a detection is considered a \ac{tp} if its \ac{iou} with the ground truth bounding box is greater than a specified threshold (e.g., 0.5), and a \ac{fp} if the \ac{iou} is below that threshold. A \ac{fn} occurs when an object in the ground truth is not detected. \Ac{ap} is then computed as the \ac{auc} of this precision-recall curve, providing a single numerical value that summarizes the model's performance for that specific class. In the \ac{voc} challenge, \ac{ap} is computed using interpolated precision at a set of 11 recall levels, which can be represented by the set \( R = \{ 0, \, 0.1, \, 0.2, \, \dots, \, 1.0 \}\). The interpolated precision \( p_{interp}(r) \) at a particular recall level \( r \in R \) is defined as the maximum precision achieved for any recall level greater than or equal to \( r \):
\begin{equation}
p_{interp}(r) = \max_{\tilde{r} \geq r} p(\tilde{r})
\end{equation}
The \ac{ap} for a particular class \( c \) is then given by:
\begin{equation}
AP_c = \frac{1}{11} \sum_{r \in R} p_{interp}(r)
\end{equation}
\adcomment{The use of 11 recall levels in this evaluation protocol is designed to strike a balance between computational efficiency and adequately representing the precision-recall curve. The 11-point interpolation smooths out small fluctuations, or "wiggles," in the precision-recall curve that arise from minor variations in the ranking of detection results. These variations are often insignificant in large evaluation datasets, and the 11-point method is sufficient to provide a stable and reliable measure of model performance.}

Complementarily, \ac{map} provides a comprehensive metric that encapsulates both precision and recall aspects of model performance by averaging the \ac{ap} values across all classes. This process involves calculating \ac{ap} for each individual class and then computing the arithmetic mean of these values.

\begin{equation}
mAP = \frac{1}{N} \sum_{i=1}^{N} AP_i
\end{equation}
where \( N \) is the number of object classes.

\adcomment{In \ac{voc}, the \ac{map}@0.5 threshold is used, where a detection is considered correct if its \ac{iou} with the ground truth is greater than 50\%. This threshold was deliberately set low to account for inaccuracies in bounding boxes in the ground truth data.}
 
As highlighted in recent literature, such as the comprehensive review of \ac{yolo} architectures by Terven and Cordova-Esparza~\cite{terven2023comprehensive}, the \ac{map} metric has evolved in tandem with advancements in \ac{od} frameworks. For instance, \ac{yolo} models typically leverage the \ac{coco} dataset for benchmarking, which employs a more sophisticated method for calculating \ac{ap}, incorporating multiple \ac{iou} thresholds. \adcomment{The introduction of 101 recall points and multiple \ac{iou} thresholds by \ac{coco} represents a natural evolution of evaluation techniques in \ac{od}. As \ac{od} models began to saturate, it became clear that models with equivalent scores were not performing equally. \ac{coco} addressed this by increasing the granularity of evaluation, with \ac{map}@[0.5:0.95], calculated for 10 different \ac{iou} thresholds ranging from 0.5 to 0.95 in 0.05 increments. This evolution provides a more detailed performance profile, allowing for more accurate comparisons between models by assessing their localization accuracy across varying levels of overlap with ground truth boxes.} \adcomment{This method provides a finer evaluation of model performance compared to the simpler 11-point interpolation method used in \ac{voc}, enabling a better understanding of detection performance under stricter localization constraints.}

\subsection{Efficiency evaluation on embedded systems}
\label{subsec:effi}
Evaluating the performance of \ac{od} models on embedded systems necessitates a comprehensive understanding of several key metrics. Each metric provides insights into how effectively the system operates under the constraints characteristic of resource-limited edge computing environments. This section outlines the essential performance indicators used in benchmarking \ac{od} models deployed on embedded systems. Some of these metrics are \textbf{target-independent}, providing insights that apply regardless of the specific hardware used, while others are \textbf{target-dependent}, as their values can vary depending on the capabilities and constraints of the hardware platform.

\noindent \textbf{Computational complexity (Target-independent).} \Acp{flop} and \acp{mac} are crucial metrics for evaluating the computational complexity of \ac{od} models. \acp{flop} provide a measure of the total number of floating-point operations required encompassing additions, multiplications, and other arithmetic operations while \acp{mac}, a specific type of operation, represent the combined multiplication and addition operations frequently encountered in \acp{nn} (especially in convolutional layers and matrix multiplications), often serving as a proxy for overall computational complexity. Both \acp{flop} and \acp{mac} directly correlate with the processing power and computational demands of a model. \adcomment{It is important to note that while \acp{flop} and \acp{mac} are target-independent measures (quantifying the total number of operations performed by a model), their practical impact on runtime and energy efficiency depends on hardware-specific factors. Efficiency is often associated solely with metrics such as \acp{flop}/W (operations per second per Watt), but relying exclusively on these measures can be misleading. In practice, actual performance is also governed by factors like parallelism, memory bandwidth, and data movement overhead which these metrics do not capture~\cite{sze}).} \adcomment{For clarity, these metrics are typically expressed in MegaFLOPs (\(10^6\) \acp{flop}) and MegaMACs (\(10^6\) \acp{mac}), providing a theoretical measure of a model's computational load.} Benchmarking models on these metrics aids in selecting architectures that balance accuracy with computational efficiency. Reference models such as MobileNets~\cite{MobileNetV1} and ShuffleNet~\cite{ShuffleNet} are designed to significantly reduce both \acp{flop} and \acp{mac} while maintaining robust performance, making them well-suited for resource-constrained environments.

\noindent \textbf{Memory Footprint (Target-dependent).} Given the stringent memory constraints often imposed on embedded systems, optimizing the memory footprint of \ac{od} models is of paramount importance. Efficient utilization of \ac{sram} ensures the system can effectively manage its resources to perform tasks without necessitating additional hardware enhancements. \adcomment{The unit for memory usage is typically in kilobytes (KB) in \ac{tinyml} applications, as these are common for low-power, edge devices with limited memory resources.} Techniques for reducing memory requirements include model compression and the adoption of streamlined \acp{nn} architectures. Notably, evaluations of models like MCUNetV1~\cite{mcunet}, MCUNetV2~\cite{mcunetv2}, EtinyNet-\acs{ssd}~\cite{Xu_Li_Zhang_Lai_Gu_2022}, TinyissimoYOLO~\cite{moosmann2023flexible}, and XiNet~\cite{ancilotto2023} have demonstrated significant progress in this area, showcasing ultra-efficient designs that maintain high performance even within the tight memory limitations characteristic of \acp{mcu}. These models will be discussed in more detail in Section~\ref{sec:object-detection-embed-app}.

\noindent \textbf{Runtime (Target-dependent).} Runtime is an important metric, especially for real-time applications such as autonomous navigation and interactive systems. It is crucial to minimize response times without compromising accuracy to ensure the system remains both reliable and efficient under operational conditions. \adcomment{The unit of runtime is typically in seconds (s) or milliseconds (ms)}. Research endeavors, such as those presented by Han Cai {\it et al.}~\cite{once-for-all} on dynamically adjustable networks, demonstrate innovative methodologies for optimizing runtime while preserving high levels of accuracy.

\noindent \textbf{Energy Consumption (Target-dependent).} Energy efficiency is of paramount importance in embedded systems, particularly those dependent on battery power or situated in remote locations. Judicious energy utilization directly translates to extended operational lifespan and enhanced system reliability.  \adcomment{The unit for energy consumption in \ac{tinyml} applications is typically in milliJoules (mJ) or microJoules (\textmu J), as these are standard for measuring small-scale power consumption in embedded devices.} Benchmarking energy consumption, as shown in the research by Moosmann {\it et al.} with TinyissimoYOLO~\cite{moosmann2023flexible}, involves a meticulous assessment of how model adaptations can curtail power usage while preserving functional integrity.
\begin{table}[t]
\centering
\captionsetup{justification=centering} 
\caption{\adcomment{Comparison of TinyML Hardware Platforms}}
\label{tab:tinyml_comparison}
\resizebox{0.95\columnwidth}{!}{
\begin{tabular}{lccccccc}
\toprule
\textbf{Platform} & \textbf{Core} & \textbf{Frequency} & \textbf{SRAM} & \textbf{Flash} & \textbf{AI Accelerator} & \textbf{Approx. Cost} \\
\midrule
STM32H743      & Cortex-M7 & 480 MHz & 1 MB    & 2 MB           & \xmark        & \$5--\$10  \\
STM32F746      & Cortex-M7 & 216 MHz & 320 KB  & 1 MB           & \xmark        & \$4--\$8   \\
STM32N6        & Cortex-M55 + NPU & 800 MHz & 4.2 MB  & 8MB (external) & \ \adcomment{\checkmark} & \$20--\$25 \\
GAP9           & 8-core RISC-V cluster & 175 MHz & 1.6 MB  & 2 MB           & \ \adcomment{\checkmark}  & \$10--\$15 \\
Kendryte K210  & Dual-core RISC-V (RV64GC) & 400 MHz  & 512 KB  & 8 MB (external)  & \ \adcomment{\checkmark}  & \$5--\$10  \\
MAX78000       & Cortex-M4F & 100 MHz & 128 KB  & 512 KB         & \ \adcomment{\checkmark}  & \$10--\$15 \\
ESP32          & Dual-core Tensilica LX6 & 240 MHz & 520 KB  & 4 MB (external) & \xmark       & \$2--\$4   \\
Himax WE-I Plus& ARC EM9D DSP & 400 MHz & 2 MB    & 2 MB           &  \xmark       & \$10--\$20 \\
\bottomrule
\end{tabular}
}
\end{table}
\section{\adcomment{TinyML Hardware Platforms}}
\label{sec:tinyml-hardware}
\adcomment{\Ac{tinyml} hardware platforms are essential for enabling the deployment of \ac{dl} models in resource-constrained environments. Unlike cloud-based architectures, TinyML platforms must efficiently balance computational throughput, memory constraints, and energy consumption to achieve real-time inference. These platforms mainly consist of \acp{mcu} with limited resources, though some integrate specialized \ac{ai} accelerators to enhance efficiency for \ac{od} tasks.} \adcomment{Table~\ref{tab:tinyml_comparison} presents a comparative overview of key \ac{tinyml} hardware platforms. General-purpose \acp{mcu}, such as the STM32H743, integrate a Cortex-M7 core running at 480~MHz with 1~MB of \ac{sram} and 2~MB of Flash, providing sufficient processing power for lightweight \acp{nn}. The STM32F746, featuring a lower clock speed of 216~MHz and reduced memory capacity (320~KB \ac{sram}), prioritizes power efficiency. Recent advancements, such as the STM32N6 series, introduce a Cortex-M55 core with an integrated \ac{npu}, leveraging hardware acceleration to optimize inference performance while maintaining stringent power constraints. Ultra-low-power platforms such as the ESP32 and Himax WE-I Plus further expand \ac{tinyml} applications to energy-sensitive environments.} \adcomment{Beyond traditional \acp{mcu}, some platforms incorporate dedicated \ac{ai} accelerators to enable low-power, high-performance inference. For instance, the MAX78000 combines a Cortex-M4F core with an embedded \ac{cnn} accelerator, facilitating efficient low-power \ac{od}. Similarly, the Kendryte K210, a dual-core RISC-V processor with an integrated Kendryte Processing Unit (KPU), is specifically optimized for \acp{cnn}, providing an efficient balance between power consumption and performance.} \adcomment{Among advanced \ac{tinyml} accelerators, the GreenWaves GAP9 stands out as a highly efficient platform capable of executing complex \ac{dl} workloads while maintaining low power consumption. Unlike traditional \acp{mcu}, GAP9 features an 8-core processing cluster, 1.6MB of \ac{sram}, and a dedicated neural accelerator, enabling real-time execution of models such as TinyissimoYOLO for smart glasses~\cite{moosmann2023ultraefficient}. Its energy-efficient design, coupled with optimized memory hierarchies and parallel processing, makes it particularly suited for \ac{od} applications requiring real-time inference in wearable and embedded devices.}

\adcomment{The acceleration of \ac{od} on \ac{tinyml} hardware is fundamentally dependent on specialized processing architectures designed to optimize key mathematical operations. Traditional \acp{mcu}, which rely solely on scalar CPU cores, face inherent limitations in executing tensor-intensive computations. To address this, modern \ac{tinyml} hardware integrates \acp{npu} and \ac{cnn} accelerators, which leverage parallelism and optimized execution pipelines to enhance performance. These accelerators significantly improve the efficiency of operations such as matrix multiplications, convolutions, and activation functions, which are computationally demanding on resource-constrained devices.}

\adcomment{Mathematically, the computational complexity of convolutional layers in \ac{cnn}-based \ac{od} models is expressed as:}
\begin{table}[t]
\centering
\resizebox{0.9\columnwidth}{!}{
\begin{threeparttable}
\captionsetup{justification=centering} 
\caption{Summary of recent lightweight \ac{od} models and their performance on \ac{coco} and \ac{voc} datasets. All results are compiled from the corresponding papers, and readers are encouraged to consult these papers for a comprehensive understanding of various hyperparameters and specific model details.}
\begin{tabular}{l l l l c c c}
\toprule
\textbf{Optim.} & \textbf{Model} & \textbf{Year} & \textbf{Optim. parts} & \textbf{Params (M)$\downarrow$ } & \textbf{AP50:95 (COCO)}$\uparrow$  & \textbf{\acs{map} (Pascal VOC)}$\uparrow$  \\
\midrule
\multirow{8}{*}{\textbf{Model Design}}
    & MobileNetV1 + SSDLite~\cite{ssd-lite} & 2018 & backbone & 5.1 & 22.2 & 68.0 \\
    & MobileNetV2 + SSDLite~\cite{ssd-lite} & 2018 & backbone & 4.3 & 22.1 & 70.9 \\
    & Tiny-DSOD~\cite{tiny-DSOD} & 2018 & backbone, neck & 1.15 & 23.2 & 72.1 \\
    & LightDet~\cite{Tang} & 2020 & backbone, neck, head & - & 24.0 & 75.5 \\
    & GnetDet~\cite{gnetdet-optimization} & 2021 & backbone, head & - & - & 66.0 \\
    & YOLOv5-N~\cite{glenn_jocher_2022_7347926} & 2022 & backbone, neck, head & 1.9 & 28.0 & - \\
    & YOLOv5-S~\cite{glenn_jocher_2022_7347926} & 2022 & backbone, neck, head & 7.2 & 37.4 & - \\
    & MobileDenseNet~\cite{MobileDenseNet} & 2023 & backbone & 5.8 & 24.8 & 76.8 \\
    & YOLOv11-N~\cite{yolo11_ultralytics} & 2024 & backbone, neck, head & 2.6 & 39.5 & - \\    
    & YOLOv12-N~\cite{tian2025yolov12} & 2025 & backbone, neck, head & 2.6 & 40.6 & - \\
\midrule
\multirow{1}{*}{\textbf{Pruning}}
    & Lite-YOLOv3~\cite{yolo-lite} & 2023 & backbone, neck, head  & 15.3 & 52.4 & 74.1 \\
\midrule
\multirow{10}{*}{\textbf{BNNs}}
    & Faster-RCNN-Bi-Real Net~\cite{Bi-Real} & 2018 & backbone & 20.1 & - & 58.2 \\
    & Unified network~\cite{Fast-objct-det} & 2018 & backbone & - & - & 44.3 \\
    & ResNet-50 GroupNet-c~\cite{structured-bin} & 2019 & backbone & - & 33.9 & 74.4 \\
    & ResNet-34 GroupNet-c~\cite{structured-bin} & 2019 & backbone & - & 32.8 & 69.3 \\
    & ResNet-18 GroupNet-c~\cite{structured-bin} & 2019 & backbone & - & 30.1 & 63.6 \\
    & Faster R-CNN-BiDet-Resnet18~\cite{Wang_2020_CVPR} & 2020 & backbone, head & 20 & 15.7 & 59.5 \\
    & Faster R-CNN-DA-BNN~\cite{ZHAO2022239} & 2020 & backbone, head  & - & - & 64.0 \\
    & Faster-RCNN-XNOR-Net~\cite{ZHAO2022239} & 2022 & backbone, head & 22.2 & - & 48.9 \\
\midrule
\multirow{9}{*}{\textbf{\ac{nas}}}
    & DetNASNet (FPN)~\cite{Yukang-detnas} & 2019 & backbone & - & 36.6 & 81.5 \\
    & DetNASNet (RetinaNet)~\cite{Yukang-detnas} & 2019 & backbone & - & 33.3 & 80.1 \\
    & NATS-C~\cite{Junran-nas} & 2019 & backbone & - & 38.4 & - \\
    & NAS-FCOS~\cite{Ning-nas} & 2019 & neck, head & - & 46.1 & - \\
    & NAS-FPNLite + MobileNetV2~\cite{Golnaz-nas} & 2019 & neck & 2.16 & 24.2 & - \\
    & MobileNetV2 + MnasFPN~\cite{Bo-nas} & 2019 & head & 1.29 & 23.8 & - \\
    & Hit-Detector~\cite{Jianyuan-nas} & 2020 & backbone, neck, head & 27 & 41.4 & - \\
    & YOLO-NAS-Small~\cite{yolo-nas} & 2021 & backbone, neck, head & 19 & 47.5 & - \\
    & MCUNetV2\tnote{$\dagger$}~\cite{mcunetv2} & 2021 & backbone & 0.67 & - & 68.3 \\
\midrule
\multirow{4}{*}{\textbf{\ac{kd}}}
    & DarkNet-YOLO-BWN-KT~\cite{Xu-al} & 2018 & backbone & - & - & 65.0 \\
    & MOBILENET-Yolo--BWN-KT~\cite{Xu-al} & 2018 & backbone & - & - & 63.0 \\
    & Centernet-34~\cite{Kuang-et-al} & 2021 & backbone, head & - & - & 75.8 \\
    & Centernet-50~\cite{Kuang-et-al} & 2021 & backbone, head & - & - & 77.1 \\
\midrule
\multirow{2}{*}{\textbf{Quantization}} 
    & EtinyNet\tnote{$\dagger$}~\cite{Xu_Li_Zhang_Lai_Gu_2022} & 2022 & backbone & 0.59 & - & 56.4 \\
    & TinyissimoYOLO\tnote{$\dagger$}~\cite{moosmann2023flexible} & 2023 & backbone, neck, head & 0.703 & - & 56.4 \\
\bottomrule
\end{tabular}
\label{benchmark-table}
\begin{tablenotes}
\item[$\dagger$] These models represent significant advancements in state-of-the-art, having been deployed on \acp{mcu}, demonstrating exceptional efficiency in memory-constrained environments.
\end{tablenotes} 
\end{threeparttable} 
}
\end{table}

\begin{equation}
    C = O_{h} O_{w} C_{o} K_h K_w C_i,
\end{equation}

\adcomment{where \( O_h, O_w \) denote the output feature map dimensions, \( C_o \) is the number of output channels, \( K_h, K_w \) are the kernel dimensions, and \( C_i \) represents the number of input channels. The presence of dedicated \ac{cnn} accelerators, such as in the MAX78000 and Kendryte K210, reduces this complexity by executing convolution operations in parallel, thereby lowering computational latency from \( O(n^2) \) to \( O(n) \) per pixel in optimized hardware.}

\adcomment{To further optimize execution, many \ac{tinyml} hardware platforms incorporate specialized memory hierarchies and direct memory access (DMA) mechanisms to facilitate efficient data movement. For example, the GAP9 integrates a tightly coupled memory system that minimizes data transfer overhead between its multi-core cluster and neural accelerator, reducing inference latency. Efficient scheduling of memory access patterns also allows for pipelined execution, improving overall throughput.} \adcomment{Further efficiency gains are achieved through specialized instruction sets tailored for \ac{dl} inference. Many \acp{npu} and \ac{cnn} accelerators support vectorized operations, which allow simultaneous execution of \ac{mac} operations. For example, the STM32N6 leverages mixed-precision arithmetic, dynamically adjusting computation precision between 16-bit and 8-bit representations to optimize both accuracy and power efficiency. This approach is particularly beneficial in \ac{od} tasks, where object localization computations often require higher precision than classification.} \adcomment{As summarized in Table~\ref{tab:tinyml_comparison}, the selection of \ac{tinyml} hardware significantly influences the feasibility of deploying \ac{od} models in real-world applications. While high-performance \acp{mcu}, such as the STM32H743, provide the computational capacity to execute \ac{od} algorithms, they lack dedicated acceleration hardware. In contrast, platforms with integrated \acp{npu} and \ac{cnn} accelerators, such as the STM32N6, GAP9, and Kendryte K210, demonstrate substantial efficiency improvements by offloading key processing tasks to specialized neural units. These advancements underscore the increasing importance of hardware-driven optimizations in achieving real-time, energy-efficient \ac{od} in \ac{tinyml} environments~\cite{moosmann2023ultraefficient}.}

\section{Advanced Techniques in Model Optimization for Embedded Systems}
\label{sec:object-detection-optim}
\begin{figure}[t]
    \centering
    \resizebox{0.85\textwidth}{!}{\input{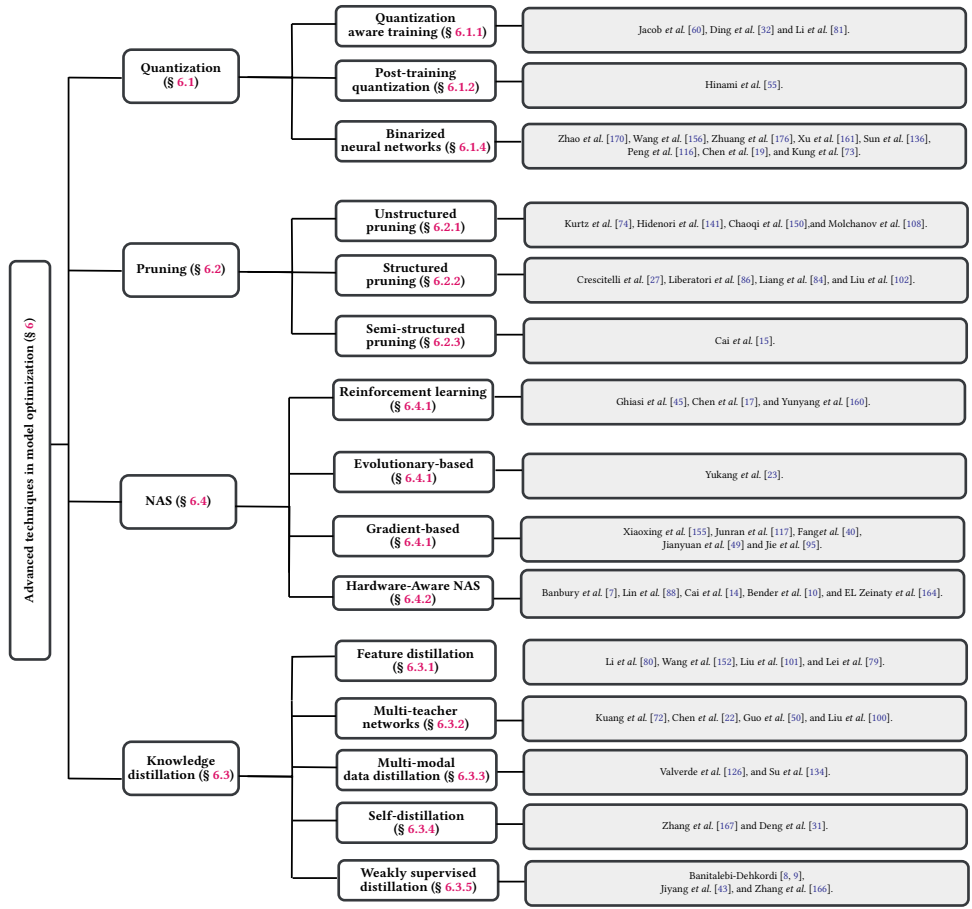}} 
    \caption{Taxonomy of model compression methods.}
    \label{fig:model_optimization}\vspace{-5mm}
\end{figure}

In this section, advanced techniques for optimizing \ac{od} models specifically designed for resource-constrained embedded systems are examined. These techniques are critical for enhancing model efficiency, reducing computational overhead, and ensuring high performance within the limited resources available in embedded environments. \adcomment{Unlike general reviews on network lightweighting, this section focuses specifically on \ac{od} as an illustrative application, highlighting how various components of the \ac{od} pipeline, namely the backbone, neck, and head, are optimized in the literature to meet the constraints of embedded \ac{ai}. By structuring the discussion around these core components, we emphasize how different optimization techniques are applied at each stage of the detection pipeline, demonstrating their impact on both computational complexity and accuracy.} Key optimization strategies are explored, broadly categorized into three primary approaches: parameter removal, parameter quantization, and parameter search. Parameter removal techniques, including unstructured, structured, and semi-structured pruning, focus on strategically eliminating redundant parameters to streamline the model and reduce its computational footprint. Parameter quantization methods, such as \ac{qat}, \ac{ptq}, and \acp{bnn}, aim to reduce the precision of model parameters, thereby lowering memory requirements and computational demands. Finally, parameter search approaches, including \ac{kd}, \ac{nas}, and its various methodologies such as \ac{rl}-based \ac{nas}, \ac{ea}-based \ac{nas}, and gradient-based \ac{nas} are discussed. Table~\ref{benchmark-table} compares some of the state-of-the-art \ac{od} models that leverage these optimization techniques, highlighting their performance on both the \ac{coco} and \ac{voc} datasets. The table illustrates the trade-offs between model complexity, parameter size, and detection accuracy (\ac{ap}50:95 and \ac{map}), showcasing how each optimization strategy impacts model performance across different benchmarks. These advanced techniques are further summarized in Fig.~\ref{fig:model_optimization}, providing a comprehensive overview of the approaches examined in this section.

\subsection{Quantization}
\label{sec:quantization}
Quantization is crucial for optimizing \acp{nn} for hardware deployment, involves converting high-precision values into lower-bit representations. This process, essential for lightweight \ac{dl}, is mathematically expressed as:
\begin{equation}
    \hat{x} = Q(x) = \text{round}\left(\frac{x}{s}\right) \times s
\end{equation}
where \( \hat{x} \) represents the quantized value of the the original value \( x \), and \( s \) the scaling factor. Quantization effectively reduces \ac{mac} operations and \ac{dnn} size, thereby accelerating both training and inference phases. Depending on whether quantization errors ($|x-\hat{x}|_p$) are taken into account during the training phase, we can distinguish between \ac{qat} and \ac{ptq}.

\subsubsection{\ac{qat}}
\label{sec:qat}
\Ac{qat} was initially introduced to mitigate the potential accuracy loss associated with quantization. This technique seamlessly integrates quantization into the training cycle itself, allowing the model to adapt and learn to minimize quantization error, defined as:
\begin{equation}
    e = |x - \hat{x}|_p
\end{equation}
where $| \cdot |_p$ represents the p-norm, quantifying the difference between the original value  \( x \) and its quantized counterpart \( \hat{x} \). By accounting for quantization effects during both forward and backward passes, \ac{qat} enables the model to adjust its parameters and learn representations that are robust to the lower precision representation.

\subsubsection{\ac{ptq}}
\label{sec:ptq} 
As an alternative to \ac{qat}, \ac{ptq} is often preferred due to its straightforward implementation. Applied after the model training process is completed, \ac{ptq} involves quantizing the weights and activations of a pre-trained model. Although easier to implement, \ac{ptq} may not achieve the same level of accuracy as \ac{qat}, as the model is not retrained to adapt to the quantization process.
\subsubsection{\adcomment{PTQ and QAT of OD models}}
Recent studies have demonstrated the effectiveness of quantization techniques in optimizing \ac{od} for various applications. For instance, Jacob {\it et al.}~\cite{Benoit} applied quantization to both weights and activations in \ac{od} models, using 8-bit integers to achieve efficient inference. Their evaluation on the \ac{coco} dataset showed that the quantized \ac{ssd}-MobileNet model retained an \ac{ap}50:90 comparable to its floating-point counterpart, with only a slight decrease in accuracy. Furthermore, the quantized model significantly reduced latency, making it well-suited for real-time \ac{od} applications on resource-constrained devices. Similarly, Hinami {\it et al.}~\cite{Hinami} introduced \ac{caq} within a large-scale \ac{r-cnn} framework to optimize \ac{od} at scale, particularly under limited computational resources. By integrating \ac{rvq} and inverted indexing, their approach achieved notable speed-ups and memory reductions while maintaining high detection precision, this study underscores the potential of adaptive quantization techniques. Additionally, Ding {\it et al.}~\cite{Ding} proposed the REQ-\ac{yolo} framework, which utilizes heterogeneous weight quantization and the \ac{admm} for \ac{fpga}-based \ac{od}. This method exploits block-circulant matrices to improve weight compression and storage efficiency, highlighting how specialized hardware and advanced quantization can be combined to optimize \ac{od} performance, reducing both computational load and latency. Moreover, Li {\it et al.}~\cite{Rundong} developed a \ac{fqn} tailored for \ac{od}, leveraging low-bit arithmetic for enhanced computational efficiency. Their method quantizes both network weights and activations and was evaluated using a 4-bit RetinaNet detector~\cite{retinanet} with a MobileNetV2 backbone~\cite{mobilenetv2}. The results showed only a 2.0\% \ac{ap}50:95 loss compared to the full 32-bit floating-point model, showcasing the practicality and effectiveness of low-bit models in constrained environments.
\subsubsection{Binarized Neural Networks for Object Detection}
\label{sec:bnn}
\Acp{bnn} represent an extreme form of quantization, where both weights and activations are constrained to binary values (\num{-1} or \num{+1}). This binarization process can be mathematically expressed using the sign function:
\begin{equation}
B(x) = sign(x) = 
\begin{cases}
+1, & \text{if } x \geq 0 \\
-1, & \text{if } x < 0
\end{cases}
\end{equation}

To achieve binary activations, the sign function converts floating-point values to their binary counterparts. However, this leads to uniformly zero gradients during training, posing challenges in training \acp{bnn} effectively. The \ac{ste} technique, introduced by Bengio {\it et al.}~\cite{bengio2013estimating}, is employed to facilitate effective training by approximating the gradient of the sign function.
\subsubsection{\adcomment{Binarized OD models}}
Recent research in \acp{bnn} for \ac{od} demonstrates their potential in various applications. Peng {\it et al.}~\cite{PENG201991}  propose a greedy layer-wise training method for \acp{bnn}, significantly boosting performance in detection tasks. Wang {\it et al.}~\cite{Wang_2020_CVPR} introduce a method called BiDet to enhance detection precision in \acp{bnn} through redundancy removal. Another study~\cite{Fast-objct-det} presents a fast \ac{od} algorithm using binary deep \acp{cnn}, achieving rapid detection with minimal precision loss. Further, innovative methods like Group-Net~\cite{structured-bin} and DA-BNN~\cite{ZHAO2022239} have been proposed to enhance feature representation capacity in  \acp{bnn}. Further, BNAS~\cite{BNAS} introduces binarized neural architecture search to produce highly compressed models suitable for edge computing. Kung {\it et al.}~\cite{Kung-al} and Xu {\it et al.}~\cite{Xu-al} explore the application of \acp{bnn} in infrared human detection and autonomous driving, respectively, showcasing the benefits of \acp{bnn} in terms of computational efficiency and memory usage. These studies collectively highlight the advancements and diverse applications of \acp{bnn} in \ac{od}, underscoring their efficiency and potential in resource-constrained environments.
\subsection{Network Pruning}
\label{sec:prunning}
Pruning, in the context of \ac{od} models, is a strategic approach for reducing the model's size and computational requirements. This is achieved by strategically removing or "trimming" weight parameters based on predetermined criteria. Consider a deep \ac{cnn} comprising \( L_n \) layers, where \ac{conv} layers are typically the most computationally intensive. Each layer within such a network consists of \( K_n \) kernels (or filters) and \( W_n \) non-zero weights. Consequently, the computational cost incurred during inference is directly proportional to \( W_n \times K_n \times L_n \). As \ac{dl} models grow in complexity, with increasing numbers of layers and parameters, this computational cost escalates significantly.

By implementing parameter pruning, sparsity is introduced into the model, effectively reducing the number of non-zero weights \( W_n \). Similarly, kernel pruning diminishes the number of kernels \( K_n \). These reductions in both parameters and kernels lead to a decreased overall computational burden. Modern computing platforms utilize software compression techniques that efficiently compress input and weight matrices, specifically targeting the zero-valued (pruned) parameters, allowing them to be bypassed during execution. Alternatively, specialized hardware~\cite{sonic} can directly execute these skipping operations, further enhancing efficiency. Existing pruning methodologies can be broadly classified into three categories: unstructured, structured, and semi-structured (or pattern-based) pruning, each offering distinct advantages for model optimization.
\subsubsection{Unstructured Pruning}
\label{sec:unstructured-prunning}
In unstructured pruning, weights are selectively removed from the network to minimize the impact on the loss function while preserving model accuracy. Several schemes have been developed for this purpose, including, weight magnitude pruning, gradient magnitude pruning, synaptic flow pruning, and second-order derivative pruning. The weight magnitude pruning approach zeroes out weights whose absolute values fall below a predefined threshold~\cite{pmlr-v119-kurtz20a}. Gradient magnitude pruning method targets weights with minimal gradients, as these weights are assumed to have less impact on the final output~\cite{Molchanov-cvpr}, while synaptic flow pruning attractively eliminates weights based on a global score that combines information from both weight magnitudes and their contribution to the loss function~\cite{Hidenori}. Finally, second-order derivative pruning aims to maintain network loss close to its original value by selectively zeroing out weights based on their second-order derivatives~\cite{Chaoqi}. Each of these unstructured pruning techniques offers unique advantages in terms of balancing model size reduction with accuracy preservation.
\subsubsection{Structured Pruning}
\label{sec:structured-prunning}
Structured pruning enhances both model sparsity and uniformity by removing entire filters~\cite{Crescitelli} or channels~\cite{Liu_2017_ICCV}, resulting in a more streamlined and efficient model. This uniformity not only reduces the number of \ac{mac} operations, offering an advantage over unstructured pruning, but also facilitates compatibility with hardware acceleration technologies like TensorRT~\cite{Jeong}. However, it is crucial to acknowledge that this structured approach may potentially compromise model accuracy, as it could inadvertently eliminate significant weights along with redundant ones. Nonetheless, the uniform structure of weight matrices resulting from structured pruning optimizes hardware acceleration, leading to improved memory and bandwidth utilization across various platforms.
\subsubsection{Semi-Structured Pruning}
\label{sec:semi-structured-prunning}
Semi-structured pruning, also known as pattern pruning, merges the elements of both structured and unstructured pruning. It uses kernel patterns as masks to selectively preserve weights within a kernel, thereby creating partial sparsity. The efficacy of these patterns is often assessed through metrics like the L2 norm, aiding in the identification of the most efficient masks for inference. However, this pruning method inherently induces less sparsity compared to other types due to the fixed weight reduction within kernels~\cite{Cai2020YOLObileRO}. To enhance sparsity further, it can be combined with connectivity pruning, which fully prunes specific kernels. Although this combined approach is effective, it might overlook redundant weights in 1$\times$1 kernels, as it typically targets larger kernels for more substantial pruning. Despite potential accuracy reductions stemming from the removal of critical weights, the semi-structured nature of kernel pattern pruning facilitates hardware parallelism, thereby enhancing inference speed.
\subsubsection{Pruning of OD models}
Pruning has proven to be highly effective in optimizing \ac{od} models for deployment on resource-constrained devices, as demonstrated in several recent studies. For example, Liberatori {\it et al.}~\cite{Liberatori} utilized structured pruning techniques, specifically filter pruning, to significantly enhance the performance of a \ac{yolo} model for face mask detection on low-end hardware. By pruning \ac{conv} filters with low L1-norm values, they achieved a 50\% reduction in the number of parameters, which led to nearly doubling the \ac{fps} performance on a raspberry pi-4, while maintaining only a moderate decrease of around 7\% in \ac{map}. Similarly, the Edge \ac{yolo} framework~\cite{Liang} incorporates structured pruning to streamline \ac{yolo}-based models, achieving a parameter reduction of approximately 30\% and enhancing computational efficiency without compromising detection accuracy, demonstrating a slight improvment in \ac{ap}50:95 compared to the full model. Furthermore, DeepCham~\cite{Dawei} utilizes pruning at the image level by selectively processing and discarding less relevant image data before feeding it into the \ac{od} pipeline. This approach reduces the computational load on edge devices and enhances the adaptability of embedded \ac{ai} systems in various environmental contexts. By efficiently managing which data is processed, such image-level pruning can significantly benefit embedded \ac{ai} by minimizing power consumption and improving response times in real-time applications.

\subsection{Knowledge Distillation (\ac{kd})}
\label{sec:kd}
\begin{figure}[t]
 \centering
  \includegraphics[width=0.45\textwidth]{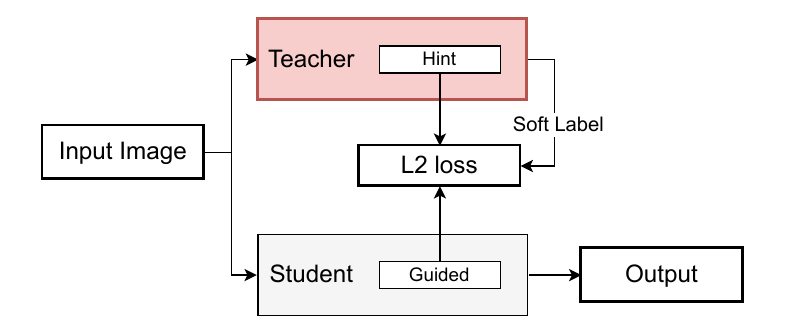}
   \caption{General hint based \ac{kd} framework.}
    \label{fig:knoledge-dist} \vspace{-5mm}
\end{figure}

\Ac{kd} is a well-established technique in the field of \ac{dl}, where a smaller, more efficient student model is trained to mimic the behavior of a larger, more complex teacher model. The teacher model's knowledge is transferred to the student model through various loss functions, enabling the student to achieve competitive performance with reduced computational resources.
FitNets~\cite{fitsnet} pioneered the extension of \ac{kd} by introducing {\it hint-based} learning, whereby the student model is trained not only to mimic the output logits of the teacher but also its intermediate feature representations, often referred to as {\it hints}. This method has markedly improved the training of student models, particularly in classification tasks, by enabling the student to leverage the richer information embedded within the teacher's internal layers. The fundamental architecture of this approach is depicted in Fig.~\ref{fig:knoledge-dist}, where the teacher network includes a hint layer, and the student network has a corresponding guided layer. The outputs from these layers, along with the soft labels provided by the teacher network, are combined to form the loss function that optimizes the student model. While this technique had demonstrated success in classification tasks, its application to \ac{od} was first explored in the seminal work by Chen {\it et al.}~\cite{NIPS2017_e1e32e23}. In this pioneering approach, the authors extended \ac{kd} to \ac{od} by incorporating three key components into the loss function, as expressed in~\eqref{eq:kd_loss2}.
\begin{equation}
L = \alpha L_{\text{det}} + \beta L_{\text{dis}} + \gamma L_{\text{hint}},
\label{eq:kd_loss2}
\end{equation}

where \( L_{\text{det}} \) represents detection-specific losses, which include both classification loss and bounding box regression loss. This component ensures the model's predictions are accurate in both object classification and localization.
\( L_{\text{dis}} \) denotes the distillation loss, where the student model is trained to mimic the soft labels (logits) produced by the teacher model and \( L_{\text{hint}} \) refers to the hint-based feature distillation loss.

This combination of losses allows the student model to leverage the teacher model's expertise in both recognizing objects and localizing them within an image. The introduction of these components marked a significant advancement in the use of \ac{kd} for more complex tasks such as \ac{od}.  To further improve the effectiveness of this framework, a variety of advanced distillation modules, mechanisms, and strategies have been developed. These include feature distillation~\cite{Li_2017_CVPR, Wang_2024_CVPR, Liu_2024_WACV, Li_feature_dist_2024}, multi-teacher networks~\cite{Chen-et-al, Kuang-et-al, Liu2019TeacherStudentsKD, Guo-kd}, multi-modal data distillation~\cite{Kruthiventi,Valverde,Su}, self-distillation~\cite{Peizhen-et-al, Deng}, and weakly supervised \ac{od} networks~\cite{Banitalebi, Dehkordi, Jiyang-et-al, CADN}, each offering unique benefits and improvements.
\subsubsection{Feature Distillation}
\label{sec:kd-feature-dist}
Although the aforementioned work extended \ac{kd} to \ac{od}, the approach primarily focused on classification and soft-label distillation, with the hint-based loss playing a supporting role. However, \ac{od} requires precise localization in addition to accurate classification, which necessitates a more targeted application of feature distillation. Li {\it et al.}~\cite{Li_2017_CVPR} propose adapting feature distillation specifically to address the localization challenges inherent in \ac{od}. Their method, termed "mimicking," involves training the student model to replicate not only the teacher's output or intermediate features in a general sense but, crucially, to focus on the region proposal features that are pivotal for \ac{od}. This approach was rigorously tested on the Faster \ac{r-cnn}~\cite{Faster-RCNN} \ac{od} framework, demonstrating substantial improvements in localization accuracy. Furthermore, the authors suggested that this methodology could be effectively applied to one-stage detectors like \ac{ssd}~\cite{ssd}, broadening its applicability across diverse \ac{od} architectures. This adaptation of the \ac{kd} loss function, focusing on feature distillation for localization, marked a pivotal shift from the earlier FitNets approach. Their work underscored that for \ac{od} tasks, where both classification and precise localization are paramount, a more targeted feature distillation strategy could yield significantly enhanced performance. While this formulation serves as a robust foundation, recent literature has presented various innovative adaptations to further refine the application of feature distillation to \ac{od}. For example, Wang {\it et al.}~\cite{Wang_2024_CVPR} introduce a cross-head knowledge distillation approach that specifically focuses on aligning predictions between the teacher and student models by sharing intermediate features between their detection heads. This method effectively addresses inconsistencies in predictions, ensuring that the student model captures both high-level semantic information and fine-grained details, thereby improving detection accuracy, particularly for small objects. Similarly,  Liu {\it et al.}~\cite{Liu_2024_WACV} extend this concept by proposing an attention-guided distillation technique, ensuring that the student network focuses on the most informative regions of the teacher's feature maps. This approach optimizes both accuracy and efficiency in complex detection scenarios. In another notable development,  Li {\it et al.}~\cite{Li_feature_dist_2024} present a hierarchical feature mimicking strategy that leverages multi-level feature distillation, enabling the student model to better understand spatial and semantic characteristics across different scales and contexts. Additionally, the authors introduce a novel approach by distilling knowledge across different heads of \ac{od} networks. This cross-head distillation technique allows the student model to more effectively leverage the diverse knowledge encoded in multiple detection heads, thereby enhancing both classification and localization performance.

\subsubsection{Multi-Teacher Networks}
\label{sec:kd-multi}
Multi-teacher networks represent an advancement in \ac{kd}, incorporating multiple teacher models, each contributing unique knowledge representations. This approach leverages the collective expertise of the teacher ensemble to provide the student network with a richer and more diverse set of features and decision boundaries. By distilling knowledge from these multiple sources, the student network gains a more comprehensive understanding of the \ac{od} task, leading to improved generalization across varied data distributions.

Chen {\it et al.}~\cite{Chen-et-al} showcase the application of multi-teacher networks in \ac{od} by introducing an enhanced knowledge distillation framework. In their approach, multiple teacher models are first trained to achieve high accuracy on the original dataset. Subsequently, a student model, structurally distinct from the teacher models in its backbone layer but sharing the same output layer, is trained using both soft targets generated by the teacher models and hard targets derived from the original dataset. This strategy enables the student model to benefit from the deep, specialized knowledge encapsulated within the multiple teachers while also ensuring alignment with the ground-truth labels. Furthermore, a weighting mechanism is incorporated, whereby the heatmap representing the size and offset for each object category is adjusted based on the classification accuracy of each teacher model. This refinement further optimizes the distillation process, facilitating more effective knowledge transfer. A similar multi-teacher approach was adopted by Kuang {\it et al.}~\cite{Kuang-et-al}, who designed a knowledge distillation framework tailored for \ac{od} using CenterNet~\cite{centernet} as the base architecture. In this approach, multiple teacher models are used to enhance the student model's learning by providing soft targets derived from the heatmaps generated by the teachers. These heatmaps are weighted and fused according to each teacher model's classification accuracy, ensuring a balanced knowledge transfer.

Alternatively, a single teacher model can guide multiple student models concurrently. Liu {\it et al.}~\cite{Liu2019TeacherStudentsKD}~investigate this approach in the context of siamese visual tracking, where a single, well-trained teacher model distills its knowledge into several student models. Each student model is designed to specialize in a particular facet of the tracking task, such as robustness to appearance changes or precise localization. By training these student models in parallel and encouraging them to focus on distinct strengths, the overall tracking performance is substantially enhanced. This method also facilitates the selection of the most suitable student model for deployment, tailored to the specific requirements of the task at hand.

\begin{figure}[t]
 \centering
  \includegraphics[width=0.45\textwidth]{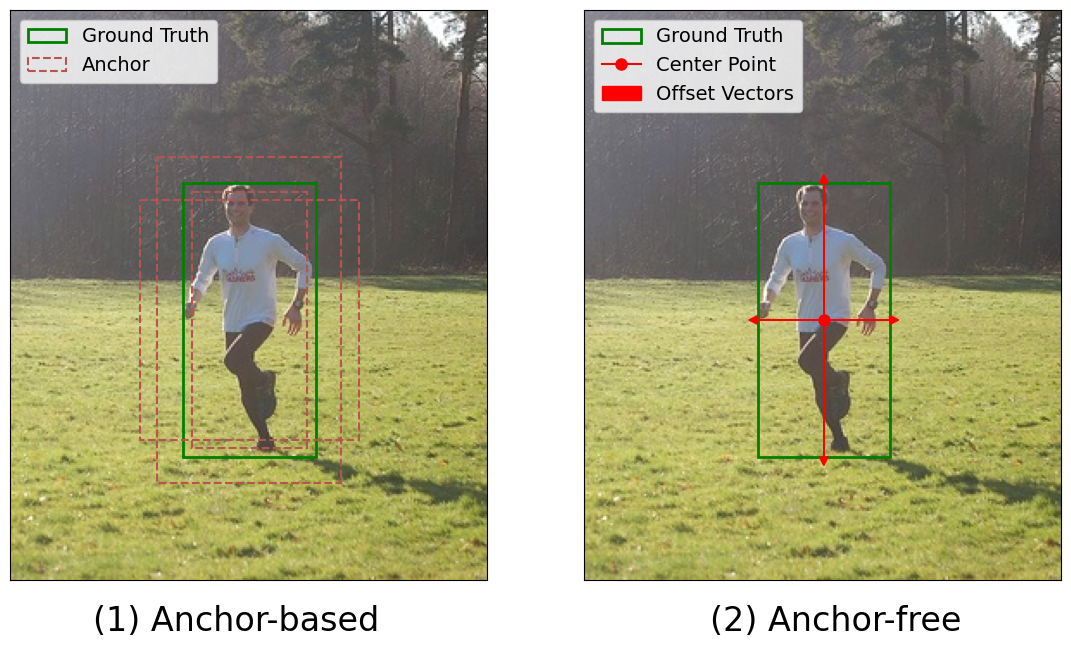}
   \caption{Comparison between anchor-based and anchor-free detection methods. In both methods, the green box is the ground-truth. In the anchor-based method (1), the red dashed boxes represent predefined anchors that guide the object detection, with no center point shown. In the anchor-free method (2), the red arrows indicate the offset distances from the predicted center point (red dot) to the edges of the bounding box. The bounding box is directly regressed in the anchor-free method, without using predefined anchors.}
    \label{fig:anchor}\vspace{-5mm}
\end{figure}

Finally, drawing inspiration from collaborative learning, an approach where students guide each other has also been investigated. In this framework, while a teacher model provides initial guidance, the student models are also designed to learn from one another, sharing knowledge to mutually enhance their performance. Specifically, the shared knowledge distillation (shared-\acs{kd}) method~\cite{Guo-kd} exemplifies this concept by enabling student models to engage in cross-layer distillation within themselves. In this approach, student models not only learn from the teacher but also mimic and refine features across their own layers, effectively bridging the semantic gap between the teacher and student networks. The experimental results presented in the study demonstrated that shared-\acs{kd} consistently improves detection performance across a diverse range of \ac{od} architectures, including two-stage, one-stage, and anchor-free detectors (Fig.~\ref{fig:anchor}). Moreover, this method significantly outperforms traditional knowledge distillation techniques, establishing it as a robust and efficient solution for real-time \ac{od} tasks.

\subsubsection{Multi-Modal Data Distillation Networks}
\label{sec:kd-multi-data}
Multi-modal data distillation networks extend the traditional \ac{kd} framework by incorporating and distilling knowledge from multiple data modalities. In contrast to unimodal approaches, where the student network learns exclusively from a single type of data (e.g., images), multi-modal data distillation leverages additional modalities such as text, depth, or audio, in conjunction with visual data. This integration of diverse data sources empowers the student network to develop a more holistic understanding of the \ac{od} task, capturing richer and more nuanced information that may not be fully represented by any single modality in isolation.

For instance,~\cite{Valverde} applied multi-modal data distillation to \ac{od} and tracking by distilling knowledge from multiple teacher networks, each trained on a distinct modality (e.g., RGB, depth, and thermal images). The student network, distilled from these diverse sources, learns to effectively integrate the complementary strengths of each modality. This enables the student network to perform robust \ac{od} even under challenging conditions where single modality data might be unreliable, such as in scenarios with poor lighting or complex backgrounds. In a similar vein,~\cite{Su} demonstrated a real-time on-road risk detection framework that utilizes multi-modal data distillation to enhance detection accuracy on mobile devices. Their approach leverages both RGB and depth images to train the teacher model, which is then distilled into a smaller, more efficient student model. This multi-modal distillation process not only improves detection performance but also ensures that the resulting model is sufficiently lightweight for deployment on resource-constrained platforms, such as mobility scooters equipped with Intel Neural Compute Stick 2 (NCS-2).

\subsubsection{Self-Distillation}
\label{sec:kd-selfdist}
Self-distillation is a sophisticated technique in which a \ac{nn} functions as both the teacher and the student, essentially distilling knowledge from its own internal representations. This process can be implemented either across different layers within the same model or over successive iterations of the model's training. In the context of \ac{od}, self-distillation serves to refine the model's internal representations and decision-making processes, thereby enhancing its capability to accurately detect and localize objects.

Peizhen {\it et al.}~\cite{Peizhen-et-al} are the pioneers in applying self-distillation to \ac{od}, introducing the \ac{lgd} framework. \Ac{lgd} eliminates the need for a separate, pre-trained teacher model. Instead, the model guides itself using label-encoded information to generate object-wise descriptors that capture both appearance and spatial characteristics. These descriptors are subsequently utilized to model inter-object relationships through a cross-attention mechanism, facilitating the network's refinement of its internal feature representations. The \ac{lgd} framework also incorporates an intra-object knowledge mapper, which maps the refined features back onto the feature map, aligning them with the spatial and geometric properties of the detected objects. This self-guided distillation process enables the model to continually enhance its detection capabilities by learning from its own predictions, offering a resource-efficient approach particularly advantageous in scenarios with limited computational resources. Building upon this concept, a recent development in the field is the \ac{sssd} framework~\cite{Deng}. \Ac{sssd} further advances self-distillation by addressing certain limitations present in earlier methods, such as those employed in \ac{lgd}. Notably, \ac{sssd} introduces the use of \ac{js} divergence~\cite{Englesson} as a smoother and more robust alternative to the traditional \ac{mse} loss. This proves particularly beneficial in the complex feature space encountered in \ac{od} tasks. In \ac{od}, where models must contend with diverse and noisy input data (e.g., occlusions, varying object scales, and cluttered backgrounds), \ac{js} divergence helps maintain stable training and improves the model's ability to generalize across different detection scenarios. Moreover, \ac{sssd} incorporates a stepwise adjustment of the distillation coefficient \(\beta\) in \eqref{eq:kd_loss2}, based on the learning rate schedule. This dynamic adjustment is crucial for sustaining effective knowledge distillation throughout the training process. As the learning rate decreases in the later stages of training, the model approaches convergence, and adjusting \(\beta\) ensures that the distillation loss remains balanced with the task-specific loss. This prevents the model from prematurely halting its knowledge transfer from the teacher model, enabling continued refinement and avoiding overfitting in the final stages.

\subsubsection{Weakly Supervised Object Detection Networks}
\label{sec:kd-weakly}
\Ac{wsod} networks are designed to function effectively with limited or incomplete annotations~\cite{Jiyang-et-al}, a prevalent scenario in large-scale data collection where obtaining fully labeled datasets is challenging or costly. These networks leverage weak supervision signals, such as image-level labels or partial bounding box annotations, to train \ac{od} models capable of generalizing well even in the presence of sparse ground-truth data. The integration of weak supervision into \ac{kd} frameworks enables the student network to learn from the available limited labeled data while simultaneously benefiting from the distilled knowledge imparted by the teacher network. This synergistic combination proves particularly potent in scenarios where acquiring fully annotated datasets is impractical or infeasible. By utilizing \ac{kd} to transfer robust knowledge from the teacher to the student, \ac{wsod} networks can attain competitive performance, often approaching that of fully supervised models, while significantly reducing the reliance on extensive manual labeling efforts~\cite{Huang_2020_CASD}. 

An illustration of this approach can be found in the work of Banitalebi-Dehkordi~\cite{Banitalebi}, who extends the concept of \ac{wsod} by leveraging unlabeled data in conjunction with \ac{kd} techniques to train low-power \ac{od} models. In their methodology, a teacher model generates pseudo-labels from unlabeled data, which are subsequently utilized to train the student model in a weakly supervised fashion. This approach effectively decouples the distillation process from the necessity of extensive ground-truth data, making it particularly suitable for scenarios where labeled data is scarce or expensive to acquire. Moreover, this framework allows for the incorporation of multiple teacher models, each potentially specializing in different object classes or utilizing diverse architectures, thereby enriching the training of the student model. The utilization of unlabeled data not only broadens the model's applicability but also enhances its robustness by incorporating a wider array of knowledge sources. Recent advancements in unsupervised domain adaptation for \ac{od} have further underscored the significant potential of \ac{wsod} frameworks. By leveraging weak supervision in conjunction with domain adaptation techniques, these frameworks enable \ac{od} models to effectively generalize across disparate domains, such as those characterized by varying lighting conditions, weather changes, or sensor differences, without requiring annotated data in the target domain~\cite{CADN}. 

\subsection{Neural Architecture Search (\ac{nas})}
\label{sec:nas}
\Ac{nas} is a sophisticated process that aims to automate the design and configuration of \ac{nn} architectures. In the field of \ac{od}, these architectures typically consist of three primary components: a backbone, a neck, and a head, as illustrated in Fig.~\ref{fig:one-two-stage-archi}. During the iterative process of \ac{nas}, the search strategy evaluates candidate architectures for each component (backbone, neck, or head) based on performance metrics such as \ac{map}. This feedback loop refines the search strategy, guiding it towards increasingly promising architectures. This automated approach significantly accelerates the design of efficient and effective \ac{od} models.
\begin{figure}[t]
 \centering
  \includegraphics[width=0.6\textwidth]{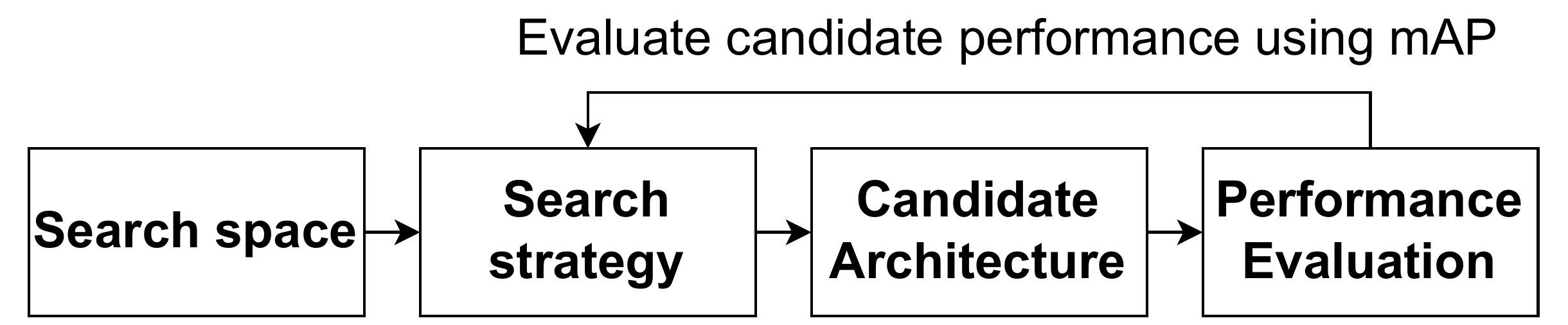}
   \caption{\Acl{nas} framework.}
    \label{fig:nas-framework}\vspace{-5mm}
\end{figure}
The core challenge in \ac{nas} lies in defining the appropriate search space and identifying the optimal \adcomment{multi-objective search strategies} to explore this space efficiently. As illustrated in Fig.~\ref{fig:nas-framework}, the general \ac{nas} framework involves an iterative process where the search space, which can focus on potential architectures for either the backbone, neck, or head, is explored through various search strategies. These strategies encompass different methodologies, including \ac{rl} and \ac{ea}, both of which played a prominent role in early \ac{nas} research~\cite{Bo-nas, Yunyang-nas, Golnaz-nas, Yukang-detnas}. While effective, these traditional methods often proved computationally expensive, frequently necessitating extensive computational resources over many \ac{gpu} days to identify optimal architectures.

To address this efficiency concern, differentiable \ac{nas}~\cite{Xiaoxing, Junran-nas, FNA-nas,Jianyuan-nas, Jie-nas} was proposed, leveraging stochastic gradient descent to guide the search process. This approach diverges from earlier methods by transforming the inherently discrete and non-differentiable search space into a continuous one. This continuous relaxation facilitates the application of gradient-based optimization, rendering the search process more efficient.

\adcomment{Beyond optimizing for accuracy, recent advancements in \ac{nas} emphasize multi-objective search, where additional constraints such as computational cost, model size, and energy efficiency are incorporated into the optimization process. This shift is particularly relevant in the context of resource-constrained environments, leading to the development of \ac{hnas}. Unlike conventional methods, \ac{hnas} explicitly integrates deployment constraints such as latency, power consumption, and memory footprint into the search framework, ensuring that the generated architectures are not only performant but also feasible for real-world deployment.}
\begin{table}[t]
\centering
\captionsetup{justification=centering} 
\caption{Comparison of \ac{nas}-based \ac{od} models evaluated on the \ac{coco} dataset. An asterisk ("*") denotes TPU-days; all other values are in \ac{gpu} days.}
\label{tab:nas_comparison_coco}
\resizebox{0.65\columnwidth}{!}{
\begin{tabular}{l l l c c}
\toprule
\textbf{Method} & \textbf{Search} & \textbf{Optim. Parts} & \textbf{AP50:95}$\uparrow$ & \textbf{Search Cost}$\downarrow$ \\
\midrule
\ac{nas}-FCOS~\cite{Ning-nas} & \ac{rl} & Neck, Head & 43.0 & 28 \\
MobileDets~\cite{Yunyang-nas} & \ac{rl} & Backbone & 28.5 & - \\
NAS-\ac{fpn}~\cite{Golnaz-nas} & \ac{rl} & Neck & 44.2 & 333* \\ \midrule
DetNAS~\cite{Yukang-detnas} & \ac{ea} & Backbone & 42.0 & 44 \\ \midrule
EAutoDet-s~\cite{Xiaoxing} & Gradient & Backbone, Neck & 40.1 & 1.4 \\
EAutoDet-m~\cite{Xiaoxing} & Gradient & Backbone, Neck & 45.2 & 2.2 \\
EAutoDet-L~\cite{Xiaoxing} & Gradient & Backbone, Neck & 47.9 & 4.5 \\
EAutoDet-X~\cite{Xiaoxing} & Gradient & Backbone, Neck & 49.2 & 22.0 \\
ResNet-101 (NATS)~\cite{Junran-nas} & Gradient & Backbone & 40.4 & 20.0 \\
ResNeXt-101 (NATS)~\cite{Junran-nas} & Gradient & Backbone & 41.6 & 20.0 \\
FNA++ (RetinaNet)~\cite{FNA-nas} & Gradient & Backbone & 34.7 & 8.5 \\
FNA++ (SSDLite)~\cite{FNA-nas} & Gradient & Backbone & 23.9 & 21.0 \\
Hit-Detector~\cite{Jianyuan-nas} & Gradient & Backbone, Neck, Head & 41.4 & - \\
IC-ResNet50~\cite{Jie-nas} & Gradient & Backbone, Neck, Head & 39.2 & - \\
\bottomrule 
\end{tabular}
}
\end{table}
Table~\ref{tab:nas_comparison_coco} presents a comparison of \ac{od} models designed using \ac{nas}, evaluated on the \ac{coco} dataset. The table summarizes key aspects such as the search methods employed \adcomment{within the multi-objective search framework} (\ac{rl}, \ac{ea}, and gradient-based approaches), the specific architectural components optimized, performance (\ac{ap}50:95), and the associated search costs. \adcomment{The following sections provide a structured discussion of these search strategies, followed by an in-depth exploration of \ac{hnas}, which extends \ac{nas} methodologies to account for real-world hardware constraints.}
\subsubsection{\adcomment{Multi-Objective Search Strategy}}
\noindent \paragraph{\textbf{Reinforcement Learning-based \ac{nas}}}
\phantomsection
\label{sec:rl}
\Ac{rl}-based \ac{nas} employs a controller, typically a \ac{rnn}, to sample architecture configurations from a predefined search space~\cite{Giles}. The controller is trained to maximize the expected validation accuracy of the generated architectures, utilizing this accuracy as the reward signal. This process is formalized through a policy gradient method, where the controller's policy \( \pi(a|s;\theta) \) generates a probability distribution over possible actions \( a \) (i.e., architecture configurations) given the current state \( s \) (representing previous architecture decisions). The objective function \( J(\theta) \) for the controller, parameterized by \( \theta \), defined as the expected reward, is formulated in~\eqref{aq:reward}.
\begin{equation}
J(\theta) = \mathbb{E}_{\pi(a|s;\theta)}[R(s,a)],
\label{aq:reward}
\end{equation}
where \( R(s,a) \) represents the reward function, which is typically the model's performance on a validation dataset in the context of \ac{nas} for \ac{od}. 

The policy gradient theorem provides a methodology for computing the gradient of \( J(\theta) \) as follows:
\begin{equation}
\nabla_{\theta}J(\theta) = \mathbb{E}_{\pi(a|s;\theta)}[\nabla_{\theta} \log \pi(a|s;\theta) R(s,a)].
\end{equation}
In practice, a \ac{rl} algorithm such as REINFORCE~\cite{reinforce} can be employed to estimate the gradient expressed in~\eqref{eq:grad}.

\begin{equation}
\nabla_{\theta}J(\theta) \approx \frac{1}{N} \sum_{i=1}^{N} \nabla_{\theta} \log \pi(a^{(i)}|s^{(i)};\theta) R(s^{(i)},a^{(i)}),
\label{eq:grad}
\end{equation}
where \( N \) is the number of architectures sampled in a batch, and \( (s^{(i)}, a^{(i)}) \) represents the \( i \)-th sampled state-action pair. 

Nevertheless, \eqref{eq:grad} can exhibit high variance in the gradient estimation, potentially leading to instability in the learning process and hindering convergence. To mitigate this issue, a baseline \( b(s^{(i)}) \) is commonly introduced. This baseline, often implemented as the running average of the rewards or an estimated value function, serves to minimize the variance of the gradient by centering the rewards. Although the baseline does not affect the bias of the gradient estimate, it helps promote stability and enhances the efficiency of the training process. Incorporating this baseline subtraction, the gradient estimation equation is modified as follows:

\begin{equation}
\nabla_{\theta}J(\theta) \approx \frac{1}{N} \sum_{i=1}^{N} \nabla_{\theta} \log \pi(a^{(i)}|s^{(i)};\theta) \left( R(s^{(i)},a^{(i)}) - b(s^{(i)}) \right),
\label{eq:grad_baseline}
\end{equation}

Furthermore, \ac{enas} was developed to address the computational challenges inherent in traditional \ac{rl}-based \ac{nas}. \ac{enas} achieves this efficiency by introducing a technique known as parameter sharing among the candidate architectures sampled during the search process. Specifically, \ac{enas} constructs a single, over-parameterized model referred to as a {\it supernet}, which encompasses all possible architectures within the predefined search space. Rather than training each sampled architecture independently, \ac{enas} allows these architectures to inherit weights from the supernet. When the controller selects a candidate architecture, it is treated as a subgraph within the supernet, inheriting its corresponding weights rather than starting from random initialization. These shared weights are then updated across multiple architecture evaluations, enabling new architectures to benefit from the training of previously sampled ones.

A notable example of \ac{rl}-based \ac{nas} is demonstrated in the \ac{nas}-FCOS framework~\cite{Ning-nas}, where a \ac{lstm}-based controller predicts the complete \ac{od} architecture. As shown in Table~\ref{tab:nas_comparison_coco}, \ac{nas}-FCOS optimizes the neck and head components (\acs{fpn} structure) and achieves a high \ac{ap}50:95 of 43.0 with a search cost of 28 \ac{gpu} days. This progressive search strategy focuses on optimizing the \ac{fpn} structure and the prediction head separately, leading to significant reductions in computational resources and time. \Ac{nas}-FCOS further speeds up training by fixing the backbone network and caching its pre-computed outputs, enabling the controller to concentrate solely on optimizing the \ac{fpn} and prediction head configurations. Additionally, to improve the reward signal's effectiveness during early training, \ac{nas}-FCOS replaces the \ac{map} with a negative loss sum as the reward. This approach stabilizes the controller's training and accelerates convergence. Overall, this application showcases the power of \ac{rl}-based \ac{nas} in discovering high-performing \ac{od} architectures that are both computationally efficient and accurate.

Another application of the \ac{rl}-based \ac{nas} strategy can be found in the MobileDets framework~\cite{Yunyang-nas}. This framework extends the principles of \ac{rl}-based \ac{nas} to optimize the backbone architecture specifically for \ac{od} applications on mobile devices. As indicated in Table~\ref{tab:nas_comparison_coco}, MobileDets achieves \ac{ap}50:95 of 28.5. While the exact search cost is not explicitly stated, the framework's utilization of the TuNAS algorithm~\cite{tunas}, which builds upon the concept of \ac{enas}, combines \ac{rl} with a parameter sharing approach to efficiently search for architectures that satisfy both accuracy and latency requirements on mobile hardware. In this framework, the controller explores a search space tailored to the computational constraints of mobile devices. The key innovation in MobileDets lies in its latency-aware design, where the reward function is adapted to incorporate latency measurements along with traditional accuracy metrics. By leveraging a supernet that encompasses all potential backbone architectures, MobileDets circumvents the need to fully retrain each sampled architecture from scratch, thereby optimizing search efficiency.

Building upon these principles, the \ac{nas}-\ac{fpn} framework~\cite{Golnaz-nas} further shows the efficacy of \ac{rl}-based \ac{nas} in refining the neck component of \ac{od} models. As demonstrated in Table~\ref{tab:nas_comparison_coco}, \ac{nas}-\ac{fpn} achieves a state-of-the-art \ac{ap}50:95 of 44.2, albeit at a search cost of 333 \ac{tpu} days. In contrast to traditional manually designed \acp{fpn}, which rely on fixed cross-scale connections, \ac{nas}-\ac{fpn} automates the discovery of the optimal feature pyramid architecture. In \ac{nas}-\ac{fpn}, the \ac{rl}-based controller explores a vast search space of potential cross-scale connections, enabling the identification of more effective and scalable \ac{fpn} structures. The search process culminates in the discovery of an optimal merging cell, which is then recursively applied to construct a deep and flexible \ac{fpn}. These results underscore the capability of \ac{rl}-based \ac{nas} methodologies to optimize not only the backbone, as demonstrated in MobileDets, but also critical components such as the neck, ultimately leading to the development of more powerful and efficient \ac{od} models.

 \paragraph{\textbf{Evolutionary Algorithm-based-\ac{nas}}}
\phantomsection
\label{sec:ea}
 \Acp{ea} apply principles inspired by biological evolution, such as selection, mutation, and crossover, to the task of optimizing \ac{nn} architectures. Within the context of \ac{nas}, \acp{ea} are employed to evolve a population of candidate architectures over successive generations, with the aim of maximizing a fitness function, typically represented by the model's accuracy on a validation dataset. The evolutionary process starts with an initial population of randomly generated architectures. In each generation, the fitness of every individual within the population, corresponding to a specific network architecture, is evaluated. The fitness function \( F \) for \ac{od} models can be defined as:
\begin{equation}
F(m) = \text{mAP}(D_{\text{val}}, m) - \lambda C(m),
\end{equation}
where \( m \) represents a model architecture within the population, \( \text{mAP}(D_{\text{val}}, m) \) denotes the \ac{map} achieved on the validation dataset \( D_{\text{val}} \), and \( C(m) \) quantifies the computational cost associated with the model, typically measured in terms of the number of parameters or \ac{flop}. The hyperparameter \( \lambda \) serves as a regularization term, controlling the trade-off between accuracy and computational cost.

The selection process favors selecting architectures with higher fitness scores to serve as parents for the subsequent generation. Crossover and mutation operators are then applied to these chosen architectures to generate offspring, forming the next generation of the population. Crossover involves combining elements from two parent architectures, while mutation introduces random modifications to an architecture's configuration. An example of a mutation operation could be mathematically expressed by~\eqref{eq:muta}.

\begin{equation}
m' = \text{mutate}(m, \mu),
\label{eq:muta}
\end{equation}
where \( m' \) is the mutated architecture, \( m \) is the original architecture, and \( \mu \) defines the mutation rate which controls the extent of changes applied.

After several generations, the population converges towards architectures that exhibit optimized trade-offs between detection performance and computational efficiency. In the domain of \ac{od}, \acp{ea} have been leveraged to discover innovative architectures that maintain high accuracy on benchmark datasets while remaining computationally feasible for deployment on embedded systems. A prominent example of applying \acp{ea} in \ac{od} is the DetNAS framework~\cite{Yukang-detnas}, which employs an \ac{ea} to search for optimal backbone architectures specifically tailored for \ac{od} tasks. As shown in Table~\ref{tab:nas_comparison_coco}, DetNAS focuses exclusively on optimizing the backbone component, achieving an \ac{ap}50:95 of 42.0 with a search cost of 44 \ac{gpu} days. The search process in DetNAS begins with an initial population of randomly generated architectures that adhere to predefined computational constraints. Each architecture undergoes a two-step evaluation: first, it is evaluated on a small subset of the training set to update batch normalization statistics, and then it undergoes a full evaluation on a validation set to determine its detection performance. Based on these evaluations, the top-performing architectures are selected as parents for the next generation. New architectures are then generated through crossover and mutation operations. Crossover combines elements from two parent architectures, while mutation introduces random changes to the architectural configuration. This evolutionary process iterates over multiple generations, progressively refining the population toward architectures that achieve a balance between high detection accuracy and low computational cost. The results, reflected in Table~\ref{tab:nas_comparison_coco}, demonstrate that DetNAS effectively utilizes evolutionary strategies to identify backbone architectures that perform competitively in \ac{od} tasks, achieving a notable \ac{ap}50:95 performance.

\paragraph{\textbf{Gradient-Based \ac{nas}}}
\phantomsection
\label{sec:gradient}
Gradient-based \ac{nas}, specifically \ac{dnas}, employs a continuous relaxation of the search space to enable efficient gradient-based optimization using methods such as \ac{sgd}. This approach starts from earlier \ac{rl} and \ac{ea}-based methods by transforming the inherently discrete and non-differentiable search space into a continuous representation. This transformation allows for the application of gradient descent to guide the search process, enhancing efficiency by directly optimizing architecture parameters and weights.

The core of gradient-based \ac{nas} involves optimizing a supernet, which is a large \ac{nn} that encompasses all possible architectures within a search space. Each possible architecture is represented as a subgraph of the supernet. This method allows for the shared use of parameters among all sub-networks, significantly reducing the computational cost compared to evaluating each architecture independently. By training this supernet, gradient-based optimization efficiently learns which architectural configurations perform best, guiding the search towards optimal designs. The optimization problem in this context can be formulated as a bi-level optimization problem:
\begin{equation}
\begin{split}
&\min_{\alpha} \mathcal{L}_{val}(w^*(\alpha), \alpha), \quad \text{with} \\   w^*(\alpha) = &\arg\min_{w} \mathcal{L}_{train}(w, \alpha),
\end{split}
\end{equation}
Here, \( \mathcal{L}_{train}\) represents the training loss, while \(\mathcal{L}_{val}\) denotes the validation loss. The architecture parameters \( \alpha \) and the network weights \( w \) undergo a nested optimization process. Initially, for a given set of fixed architecture parameters \( \alpha \), the weights \( w \) are optimized to minimize the training loss:
\begin{equation}
w \leftarrow w - \eta \nabla_{w} \mathcal{L}_{train}(w, \alpha)
\end{equation}

Subsequently, the architecture parameters \( \alpha \) are updated based on the validation loss:

\begin{equation}
\alpha \leftarrow \alpha - \eta' \nabla_{\alpha} \mathcal{L}_{val}(w^*(\alpha), \alpha)
\end{equation}

This optimization process enables the supernet to efficiently learn and update the weights and architectural parameters, facilitating the selection of optimal architectures within the search space. Building upon the foundational concept of the supernet, one-shot \ac{nas} further enhances this framework by evaluating the entire search space of architectures within a single training cycle~\cite{Jiemin-nas}. While the supernet itself allows for shared weight training across multiple sub-networks, one-shot \ac{nas} extends this capability by enabling the evaluation of all potential architectures without the need for multiple, computationally expensive training iterations. In essence, once the supernet is trained, each architecture can be rapidly assessed by activating its corresponding sub-network within the supernet, leveraging the shared weights. This approach drastically accelerates the evaluation and selection of optimal architectures, circumventing the computationally intensive process of training each candidate network individually. Formally, let  \( \mathcal{A} \) represent the set of all possible architectures within the supernet, and let \( W \) denote the weights of the supernet. The performance of a specific architecture \( a \in \mathcal{A} \) is then evaluated using ~\eqref{eq:perf}.

\begin{equation}
\text{Performance}(a) = \text{Evaluate}(a, W|_a)
\label{eq:perf}
\end{equation}
where \( W|_a \) represents the subset of supernet weights allocated to architecture \( a \).  By leveraging these shared weights, one-shot \ac{nas} provides a practical and scalable method to explore vast architecture spaces efficiently, facilitating the rapid development of high-performing \ac{nn} architectures.

The adoption of gradient-based \ac{nas} has significantly advanced the field of \ac{od}, offering more efficient optimization of \ac{nn} architectures. As shown in Table~\ref{tab:nas_comparison_coco}, the EAutoDet framework, proposed by Xiaoxing {\it et al.}~\cite{Xiaoxing}, effectively integrates differentiable \ac{nas} and one-shot \ac{nas} techniques to enhance \ac{od} performance. EAutoDet employs a differentiable search strategy that transforms the search space into a continuous domain, enabling efficient optimization via gradient descent. This approach is coupled with a one-shot \ac{nas} methodology, where a supernet is constructed to encapsulate all possible architectures. Additionally, EAutoDet introduces kernel reusing and dynamic channel refinement techniques within this supernet, reducing memory requirements and computational costs. As indicated in Table~\ref{tab:nas_comparison_coco}, by focusing on optimizing both the backbone and neck components, EAutoDet achieves state-of-the-art results in \ac{od} with significantly lower search times and resource consumption compared to previous methods, requiring only 1.4 to 22 \ac{gpu} days depending on the model variant.

Furthermore, the \ac{nats} framework~\cite{Junran-nas} offers another compelling example of the effectiveness of gradient-based \ac{nas} in \ac{od}, as reflected in Table~\ref{tab:nas_comparison_coco}. Unlike traditional approaches that design new architectures from scratch, \ac{nats} focuses on transforming existing networks, such as ResNet~\cite{resnet} and ResNeXt~\cite{resnxt}, to better suit the specific demands of \ac{od} tasks. It employs a gradient-based search strategy at the channel level, optimizing the dilation rates within convolutional layers, which primarily enhances the backbone’s ability to detect objects at multiple scales without increasing computational complexity. By capitalizing on the pre-trained weights of the original networks, \ac{nats} efficiently adapts these models for \ac{od}, resulting in improved accuracy while maintaining a search cost of around 20 \ac{gpu} days.
The fast network adaptation (FNA++) method~\cite{FNA-nas}, as presented in Table~\ref{tab:nas_comparison_coco}, harnesses the versatility of gradient-based \ac{nas} to optimize \ac{od} frameworks. FNA++ builds upon the principles of differentiable \ac{nas} by employing a parameter remapping technique, allowing for the efficient adaptation of both the architecture and parameters of a pre-trained seed network, such as MobileNetV2, specifically for the task of \ac{od}. This approach, with a particular focus on optimizing the backbone component, eliminates the necessity for extensive retraining, thereby significantly reducing computational overhead. As demonstrated in Table~\ref{tab:nas_comparison_coco}, FNA++ achieves remarkable efficiency, requiring a search cost of only 8.5 \ac{gpu} days for RetinaNet and 21 \ac{gpu} days for SSDLite, while still delivering competitive \ac{ap}50:95 scores.

Liu  {\it et al.}~\cite{Jie-nas} further showcase the potential of gradient-based \ac{nas} in enhancing \ac{od} capabilities through their \ac{iceds}. This latter introduces a novel form of dilated convolution, referred to as inception convolution, designed to optimize the effective \ac{rf} by independently varying dilation patterns along spatial axes, channels, and layers within the network. To efficiently navigate this complex search space, they propose the \ac{edo} algorithm. \Ac{edo} pre-trains a supernet encompassing all potential dilation patterns and then selects the optimal configuration for each \ac{conv} layer. While specific search costs are not explicitly reported in their work, \ac{iceds} emphasizes reducing computational overhead compared to traditional \ac{nas} methods such as DARTS~\cite{Hanxiao-nas}.

Finally, the Hit-Detector framework proposed by Guo {\it et al.}~\cite{Jianyuan-nas} offers a distinctive approach by simultaneously optimizing the backbone, neck, and head components of an object detector in an end-to-end fashion, as highlighted in Table~\ref{tab:nas_comparison_coco}. This hierarchical trinity architecture search framework ensures consistency across all components of the detection pipeline, addressing the challenge of inter-component alignment that other \ac{nas} methods might neglect. Hit-Detector introduces a multi-level search space, differentiated into sub-search spaces for each key component, and utilizes gradient-based optimization to efficiently navigate the design space. This holistic approach mitigates potential mismatches and suboptimal performance that can arise from independent component optimization, thereby enhancing the overall coherence and effectiveness of the network design. Although precise search costs are not specified, the \ac{ap}50:95 performance achieved by Hit-Detector underscores its efficacy in \ac{od} tasks.
\subsubsection{\adcomment{Hardware-Aware \ac{nas}}}
\label{subsubsec:hw_nas}
\adcomment{Recent advancements in \ac{nas} have led to the emergence of \ac{hw} optimization techniques, where models are designed not only to maximize accuracy but also to meet stringent hardware constraints. Unlike conventional \ac{nas} methods, which focus primarily on improving task performance, \ac{hnas} integrates deployment constraints, such as inference latency, power consumption, memory footprint, and computational complexity, directly into the search process. This paradigm shift is particularly critical for resource-constrained environments, such as edge devices, mobile processors, and \acp{mcu}, where efficiency is just as important as accuracy.}

\adcomment{\Ac{hnas} methods typically incorporate hardware constraints at various stages of the search pipeline. In differentiable approaches, constraints such as latency and peak memory usage are integrated into the optimization objective, enabling architectures to be selected based on both performance and efficiency metrics. For example, Micronets~\cite{micronets} extends differentiable search techniques to ensure model feasibility on low-power devices by constraining activation sparsity and computational complexity. Similarly, MCUNet~\cite{mcunet} adopts a two-stage search strategy: first, architectures are designed for optimal task performance, and then they are pruned and adapted to fit within \ac{mcu} memory limitations. In \ac{rl}-based frameworks, hardware awareness is incorporated through reward functions that penalize excessive latency or computational cost, as demonstrated in methods such as ProxylessNAS~\cite{hanproxy} and TuNAS~\cite{tunas}, which dynamically profile candidate architectures on real hardware during the search process.}

\adcomment{Recent work by EL Zeinaty et al.~\cite{elzeinaty2025tinymlframework} introduced a framework that leverages \acp{llm} for \ac{nas}, incorporating a \ac{vit}-based \ac{kd} strategy and explainability. This framework optimizes accuracy, efficiency, and memory usage in three phases: (1) the search phase, where \acp{llm} guide the exploration of architectures using Pareto optimization; (2) the full training phase, where the best candidate is fine-tuned; and (3) the \ac{kd} phase, where a pre-trained \ac{vit} model distills logits into the student model. The framework reduces search time from 12.5 days to 1.5 days (worst-case 3.5 days), demonstrating efficiency in hardware-aware architecture design. Deployed on an STM32H7 \ac{mcu} with a 320 KB \ac{sram} constraint, the models show the potential of \ac{hnas} for embedded systems. This work advances \ac{hnas} for \ac{tinyml}, paving the way for future research.}
\section{Optimizing Object Detection for TinyML}
\label{sec:object-detection-embed-app}
\begin{figure}[t]
 \centering
  \includegraphics[width=0.5\textwidth]{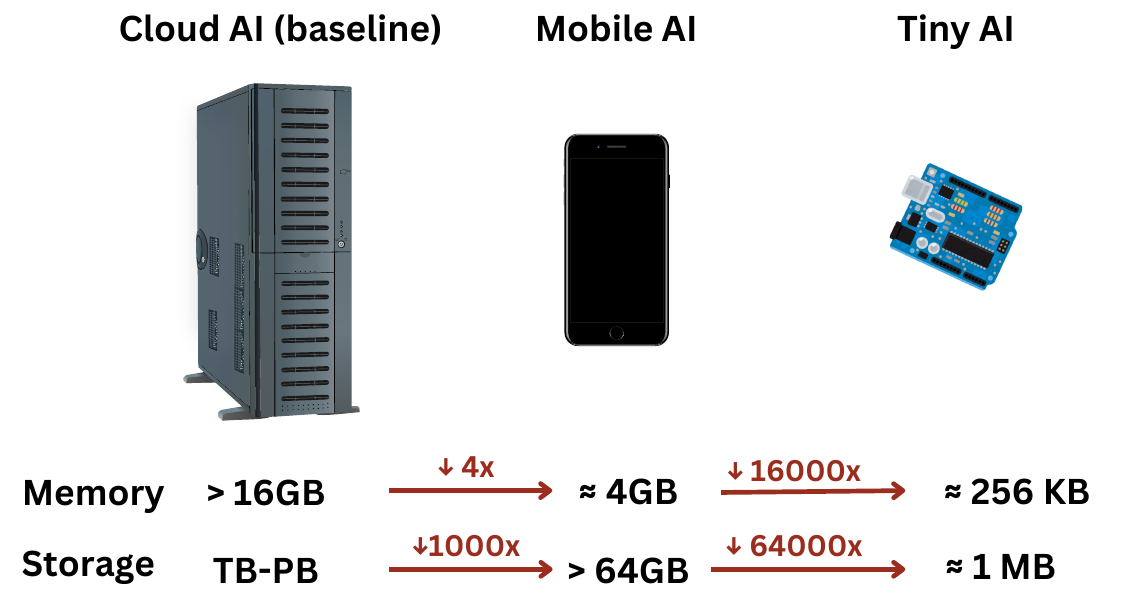}
  \caption{Comparison of computational resource constraints across different \ac{ai} platforms: cloud \ac{ai}, mobile \ac{ai}, and tiny \ac{ai}.}
  \vspace{-4mm}  
  \label{fig:edge-ai}
\end{figure}

\begin{figure}[t]
{\includegraphics[width=0.5\textwidth]{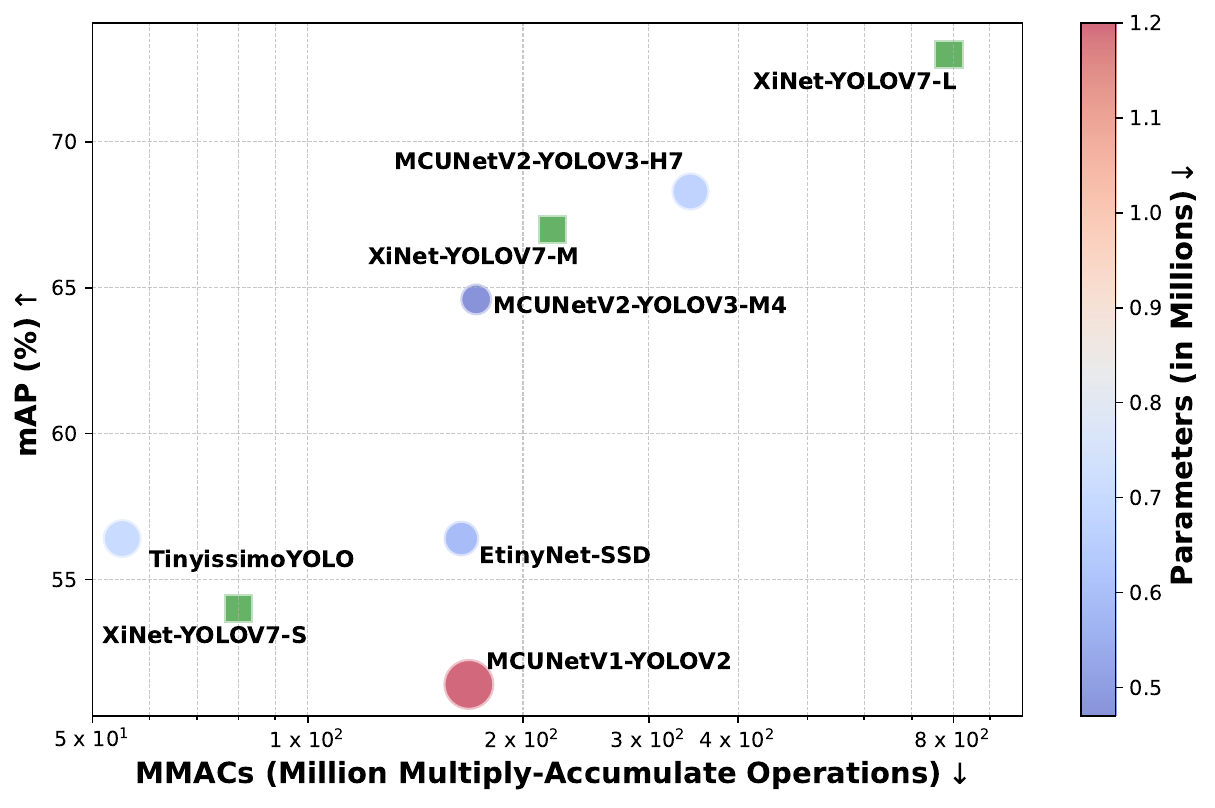}}
    \caption{Comparison of \ac{mcu}-optimized \ac{od} models in terms of MMACs, mAP, and parameters. The X-axis shows the MMACs (Million Multiply-Accumulate Operations) in log scale, representing the computational cost of each model. The Y-axis represents the \ac{map} on the \ac{voc} benchmark, indicating the accuracy of the models. The color of the bubbles corresponds to the number of parameters (in millions), with the color bar on the right indicating the scale. Larger bubbles represent models with more parameters, showing the trade-off between computational complexity and detection performance across different models. Green squares represent XiNet models, for which the parameter count was not directly specified in the paper; hence, a fixed square size was used for these models.}
    \label{fig:mmacs_map} 
      \vspace{-5mm}  
\end{figure}

In the preceding section, various optimization techniques for \ac{od}, each tailored to specific computing environments, were explored. As depicted in Fig.~\ref{fig:edge-ai}, these environments can be broadly categorized into cloud/edge \ac{ai}, mobile \ac{ai}, and tiny \ac{ai}, each characterized by distinct levels of computing power and memory resources. Cloud/edge and mobile platforms benefit from relatively abundant memory and storage capacities, with cloud/edge \ac{ai} typically exceeding 16GB of memory and boasting several terabytes of storage, while mobile \ac{ai} platforms commonly offer around 4GB of memory and upwards of 64GB of storage. In stark contrast, \ac{tinyml} operates under far more stringent constraints, typically offering between 256KB and 512KB of memory, with storage capacities in the vicinity of 1MB. \ac{tinyml} applications often run on \acp{mcu}, which are designed for low-power, low-cost scenarios~\cite{Lin-tiny-survey}. Thus, optimizing \ac{od} models for \ac{tinyml} requires not only reducing computational complexity but also ensuring that the models can fit within the restricted memory and processing capacities of \ac{mcu} platforms.
Given these extreme limitations, specialized optimization strategies are crucial for enabling \ac{tinyml} applications. Incorporating the advanced optimization techniques discussed earlier, \ac{od} models can be effectively tailored for deployment in \ac{tinyml} environments. Traditional models like \ac{yolo}, originally designed for real-time detection on \ac{gpu}-class devices, require significant modifications to operate effectively within these constrained settings. Over the years, \ac{yolo} has evolved through various iterations, offering scaled-down versions such as 'nano', 'small', 'medium', and 'large' to cater to different computational needs. However, despite these scaled-down versions, the computational demands and memory requirements of \ac{yolo} models generally remain prohibitive for \ac{tinyml} applications~\cite{aiot}.
\begin{table}[t]
\centering
\caption{Categorized Summary of Optimization Techniques Used in \ac{mcu}-Optimized Object Detection Models}
\label{table:summary_optimization}
\resizebox{\columnwidth}{!}{
\begin{tabular}{l l l}
\toprule
 \textbf{Optimization Strategy} & \textbf{Model}  & \textbf{Key Features} \\
\midrule
\multirow{3}{*}{\textbf{Neural Architecture Search (NAS)}} 
 & MCUNetV1-YOLOV2~\cite{mcunet} &   NAS with memory scheduling for MCUs; optimized memory layout reduces peak SRAM usage by 3.4$\times$. \\ \cmidrule(lr){2-3} 
 & \multirow{2}{*}{MCUNetV2-YOLOV3~\cite{mcunetv2}}  &  Use patch-based execution to reduce memory bottlenecks; in-place convolutions overwrite input   \\
 & & activations, saving 1.6$\times$ memory. \\
\midrule
\multirow{2}{*}{\textbf{Quantization-Based Optimization}} 
 & TinyissimoYOLO~\cite{moosmann2023flexible}  &   Use fully quantized (8-bit integer) models to reduce size by 4$\times$ while maintaining accuracy. \\ \cmidrule(lr){2-3} 
& EtinyNet-SSD~\cite{Xu_Li_Zhang_Lai_Gu_2022}   & Adaptive per-layer quantization and depthwise convolutions to reduce computation. \\
\midrule
\textbf{Hardware-Aware Scaling (HAS)} 
 &  XiNet-YOLOV7~\cite{ancilotto2023}  & Scale models based on RAM, flash, and compute resources; adapt to MCU memory and energy constraints.\\
\bottomrule 
\end{tabular}
}
\end{table}
While substantial progress has been made in embedded mobile \ac{ml} with models like MobileNetV1, ShuffleNet, MobileDenseNet, and Lite-YOLOv3, deploying \ac{dl} on \acp{mcu} for \ac{tinyml} presents even greater challenges. In \ac{tinyml} applications, \ac{sram} significantly restricts the size of activations (both read and write), and flash memory limits the model size, which is typically read-only. Additionally, \acp{mcu} typically operate at clock speeds ranging from 50 MHz to 500 MHz, which is considerably slower than processors found in typical laptops. \adcomment{Despite these constraints, model pruning has emerged as a crucial optimization strategy for reducing the complexity of \ac{od} models while maintaining accuracy. The DTMM framework~\cite{han2024dtmm} introduces a filter pruning technique that enables structured yet fine-grained compression, significantly reducing inference latency without compromising performance, making it particularly well-suited for \ac{tinyml} \ac{od}. Additionally, depth pruning with auxiliary networks~\cite{de2022depth} offers another promising approach, where entire layers are pruned instead of individual filters, and auxiliary networks compensate for any loss in accuracy. These methods enhance computational efficiency, allowing \acp{mcu} to run more sophisticated \ac{od} models with minimal performance degradation.
Beyond computational constraints, \ac{tinyml} \ac{od} models also face challenges related to limited training data, which can lead to overfitting and poor generalization. To address this issue, \textit{dataset distillation}~\cite{accettola} has been proposed as a technique to enable low-memory, on-device training. By synthesizing highly compact yet informative datasets, dataset distillation allows models to be efficiently trained directly on \acp{mcu}, reducing reliance on large-scale datasets and cloud-based retraining. This technique is especially beneficial for \ac{iot}-based \ac{od}, where connectivity constraints make frequent cloud updates impractical.} Innovations like TinyTracker~\cite{tiny-tracker} and the \ac{fomo} architecture~\cite{fomo2022edgeimpulse} exemplify cutting-edge approaches being developed to extend the capabilities of \ac{tinyml} in \ac{od}. TinyTracker utilizes \ac{ai} "in-sensor" technology, enabling efficient processing directly at the sensor level, thus reducing the need for data transmission and significantly lowering power consumption. Similarly, \ac{fomo}, introduced by Edge Impulse, is designed to detect and track multiple objects in real-time on highly resource-constrained devices, achieving remarkable efficiency through optimized \ac{nn} architectures and lightweight inference techniques. While TinyTracker and \ac{fomo} showcase the ongoing innovation in \ac{tinyml}, other models have been benchmarked to formally demonstrate their effectiveness. \adcomment{Table~\ref{table:summary_optimization} provides a comprehensive overview of the optimization techniques used in \ac{mcu}-optimized \ac{od} models, categorizing key strategies such as \ac{nas}, quantization, memory-aware scheduling, and hardware-aware scaling. These techniques serve as the foundation for adapting \ac{od} architectures to \ac{tinyml}, effectively addressing the trade-offs between computational efficiency, memory usage, and detection performance under severe resource constraints.} As shown in Table~\ref{table:comparison_results}, various \ac{mcu}-optimized \ac{od} models have been evaluated on the \ac{voc} benchmark. The table provides a comparative analysis of these models, highlighting their optimization techniques, computational complexity (in terms of MMACs), memory usage, and performance (\ac{map}). The results demonstrate a wide range of trade-offs between model complexity, resource consumption, and detection accuracy. These trade-offs are further illustrated in Fig.~\ref{fig:mmacs_map}, which depicts the relationship between MMACs, \ac{map} and the number of parameters for each model. In the following sections, we analyze these solutions in detail, highlighting how they address the challenges inherent in deploying \ac{od} in TinyML applications, while maintaining efficient, real-time performance under stringent power and memory constraints.

\noindent \textbf{MCUNet-YOLO.} The development of MCUNetV1-YOLOv2~\cite{mcunet} is grounded in the MCUNetV1 framework, which integrates TinyNAS and TinyEngine to optimize \ac{dl} models for deployment on resource-constrained \acp{mcu}. TinyNAS, a two-stage \ac{nas} method, begins by refining a specialized factorized hierarchical search space designed to meet the stringent resource constraints of mobile and embedded platforms~\cite{mnastnet}. This search space structures the \ac{nn} into configurable blocks, allowing for various options such as different \ac{conv} types (regular, depthwise separable, mobile inverted bottleneck), kernel sizes (3$\times$3, 5$\times$5), squeeze-and-excitation ratios, skip connections, filter sizes, and varying numbers of layers. TinyNAS optimizes this search space specifically for \ac{tinyml} applications, adjusting parameters such as input resolution $R = \{48, 64, 80, \dots, 192, 208, 224\}$ and width multiplier $W = \{0.2, 0.3, 0.4, \dots, 1.0\}$ to cover a wide spectrum of resource constraints. This approach results in $12 \times 9 = 108$ possible search space configurations, $S = W \times R$. Each search space configuration contains $3.3 \times 10^{25}$ possible sub-networks.
To efficiently identify the best models within this vast search space, TinyNAS eschews exhaustive searches by employing a one-shot neural architecture search strategy. In this approach, a supernetwork encompassing all possible sub-networks is trained using weight sharing, enabling the rapid estimation of each sub-network's performance. Subsequently, an evolutionary search algorithm is applied to select the optimal model that strikes the best balance between accuracy and resource constraints. A key component of MCUNetV1 process is TinyEngine, a memory-efficient inference library co-designed with TinyNAS. TinyEngine is utilized during the evaluation of each sampled network to optimize memory scheduling and measure optimal memory usage. Unlike traditional inference libraries that optimize memory scheduling layer by layer, TinyEngine optimizes memory usage based on the entire network topology, achieving up to a 3.4$\times$ reduction in memory consumption and a 1.7-3.3$\times$ increase in inference speed. This tight integration between TinyNAS and TinyEngine allows for the exploration of larger search spaces and the fitting of more complex models into the limited memory of \acp{mcu}, ensuring efficient and high-performance execution.
\begin{table}[t]
\centering
\captionsetup{justification=centering} 
\caption{Comparative analysis of MCU-optimized OD models using 8-bit precision for weights and activations on the VOC benchmark.}
\label{table:comparison_results}
\resizebox{0.85\columnwidth}{!}{
\begin{tabular}{lcccccccc}
\toprule
\textbf{Model} & \textbf{Venue (Year)} & \textbf{Optimization} & \textbf{Input Size} & \textbf{Parameters}$\downarrow$ & \textbf{\adcomment{MCU}} & \textbf{MMACs}$\downarrow$ & \textbf{Peak SRAM}$\downarrow$ & \textbf{mAP}$\uparrow$ \\
\midrule
MCUNetV1-YOLOV2~\cite{mcunet}  & NeurIPS (2020)  & \acs{nas} & 224$\times$224 & 1.20M & \adcomment{STM32H743} & 168 & 466kB & 51.4\% \\
MCUNetV2-YOLOV3-M4~\cite{mcunetv2}  & NeurIPS (2021) & \acs{nas} & 224$\times$224 & {\bf 0.47M} & \adcomment{STM32F412} & 172 & 247kB & 64.6\% \\
MCUNetV2-YOLOV3-H7~\cite{mcunetv2}   & NeurIPS (2021) & \acs{nas} & 224$\times$224 & 0.67M & \adcomment{STM32H743} & 343 & 438kB & 68.3\% \\
EtinyNet-\acs{ssd}~\cite{Xu_Li_Zhang_Lai_Gu_2022}   & AAAI (2022) & Quantization (\acs{asq}) & 256$\times$256 & 0.59M & \adcomment{STM32H743} & 164 & 395kB & 56.4\% \\
XiNet-YOLOV7-S~\cite{ancilotto2023} & ICCV (2023) & Model Design (\acs{has}) & 256$\times$256 & - & \adcomment{STM32H743} & 80 & {\bf 53kB} & 54.0\% \\
XiNet-YOLOV7-M~\cite{ancilotto2023} & ICCV (2023) & Model Design (\acs{has}) & 384$\times$384 & - & \adcomment{STM32H743} & 220 &  138kB & 67.0\% \\
XiNet-YOLOV7-L~\cite{ancilotto2023} & ICCV (2023) & Model Design (\acs{has}) & 416$\times$416 & - & \adcomment{STM32H743} & 789 & 511kB & \textbf{74.9\%} \\
TinyissimoYOLO~\cite{moosmann2023flexible}   &  IEEE Access (2024) & Quantization (\acs{qat}) & 112$\times$112 & 0.70M & GAP9 & {\bf 55} & - & 56.4\% \\
\bottomrule
\end{tabular}
}
\vspace{1mm}
\caption*{\footnotesize \adcomment{Note: Peak \ac{sram} indicates the maximum on-chip \ac{sram} usage during inference, expressed in kilobytes.}}\vspace{-4mm}
\end{table}
The MCUNetV1 framework has demonstrated state-of-the-art performance on ImageNet, achieving a top-1 accuracy of 70.7\% on off-the-shelf \acp{mcu}. Building upon these promising classification results, the MCUNet model was then employed as a backbone for a YOLOv2 detector to assess its effectiveness in \ac{od} tasks. Through this integration, MCUNetV1-YOLOv2 achieved a notable mAP score of 51.4\% on the \ac{voc} dataset.

Building upon the foundation established by the MCUNetV1 framework, MCUNetV2~\cite{mcunetv2} introduces several key innovations to further optimize \ac{od} on \acp{mcu}. A notable enhancement is the implementation of in-place depth-wise convolution, which substantially reduces peak memory usage by a factor of 1.6. This technique achieves efficiency by overwriting input activations with output activations during computation, thereby effectively managing the constrained memory resources typical of \acp{mcu}. Moreover, the MCUNetV2 framework addresses the challenge of imbalanced memory distribution across network layers with its innovative patch-based inference strategy. Unlike traditional layer-by-layer execution, which can lead to memory bottlenecks in the early stages of the network, MCUNetV2 processes memory-intensive layers in smaller patches. This approach considerably lowers memory requirements, enabling high-resolution processing to run efficiently on memory-limited devices, a crucial factor for effective \ac{od}.

To mitigate the computational overhead associated with patch-based inference, a method for redistributing the receptive field (\ac{rf}) throughout the network is proposed. By strategically decreasing the \ac{rf} in the initial layers to reduce the size of input patches and the frequency of computations, and subsequently expanding the \ac{rf} in later stages, the framework maintains high performance, particularly for large-scale \ac{od} tasks.
The co-design methodology of MCUNetV2 seamlessly integrates \ac{nas} with advanced inference scheduling, optimizing both the architecture design and the execution strategy concurrently. This joint optimization approach allows for greater adaptability to the stringent memory and latency constraints of \acp{mcu}, resulting in more efficient and robust \ac{od} capabilities for embedded applications. Similar to its predecessor, MCUNetV1, the new MCUNetV2 framework achieves even better results on ImageNet, attaining a top-1 accuracy of 71.8\%. For \ac{od} tasks, the MCUNetV2 model was utilized as the backbone for the YOLOv3 detector. The resulting MCUNetV2-YOLOv3 configurations (M4 and H7) achieved impressive \ac{map} scores of 64.6\% and 68.3\%, highlighting its exceptional performance in resource-constrained environments.
\noindent \textbf{TinyissimoYOLO.} TinyissimoYOLO~\cite{moosmann2023flexible} is a highly efficient and fully quantized multi-\ac{od} network, designed specifically for \ac{mcu} platforms with severe resource constraints. The network architecture, initially based on YOLOv1, has been carefully adapted to deliver real-time inference performance while operating within the stringent memory and computational limits typical of edge \ac{ai} systems. 

TinyissimoYOLO achieves this through a combination of network simplifications and optimizations. The model supports flexible input resolutions, with the ability to adjust the number of detection classes and first-layer kernel sizes to match specific deployment requirements. This modular approach allows TinyissimoYOLO to trade off between detection accuracy and computational complexity, offering a scalable solution for various \ac{mcu} architectures. Specifically, the network employs 8-bit quantization throughout, significantly reducing the memory footprint and model size by a factor of four when compared to traditional 32-bit models. For instance, TinyissimoYOLO's parameter count ranges from 441K to 887K parameters and from 32 to 57 MMACs, compared to the much larger YOLOv1 with 20 GMACs and 45M parameters. This compact design is achieved without sacrificing accuracy, making it ideal for deployment on resource-constrained \acp{mcu}.

\ac{qat} using the QuantLab framework~\footnote{\textcolor{pinkfootnote}{https://github.com/pulp-platform/quantlab}} is a key component of TinyissimoYOLO's optimization pipeline. This process ensures that 8-bit models retain accuracy close to their full-precision counterparts while also enabling efficient integer-only deployment. The QuantLab framework automates the conversion of full-precision models into fake-quantized and fully deployable integer models, making it easier to deploy TinyissimoYOLO on various hardware platforms, including RISC-V and ARM-based \acp{mcu}.

In experimental results, TinyissimoYOLO demonstrated its flexibility by achieving a \ac{map} score of 56.4\% on the \ac{voc} dataset, while operating with only 55 MMACs. This makes TinyissimoYOLO well-suited for real-time, ultra-low-power \ac{od} on \acp{mcu}, such as the GAP9 multi-core RISC-V processor and the MAX78000 with an \ac{ai} accelerator. Moreover, the network supports dynamic scaling for different \ac{mcu} configurations, allowing developers to choose between low-latency or high-energy efficiency modes depending on the application.

\noindent \textbf{EtinyNet-\acs{ssd}.} EtinyNet-\acs{ssd}, as detailed in~\cite{Xu_Li_Zhang_Lai_Gu_2022}, is meticulously designed for efficient \ac{od} on \acp{mcu}. It incorporates two key innovations: the \ac{dlb} and \ac{asq}. The \ac{dlb} seamlessly integrates depthwise separable convolutions with linear layers, substantially reducing the model's computational complexity while maintaining accuracy. This design ensures the network remains both lightweight and efficient, rendering it particularly suitable for deployment on \acp{mcu}. The backbone of EtinyNet is then leveraged in conjunction with the \ac{ssd} architecture to achieve superior performance in \ac{od} tasks. \Ac{asq} further enhances EtinyNet-\acs{ssd} by dynamically adjusting quantization scales for both weights and activations based on the input data. This technique optimizes the precision of quantized parameters, thereby minimizing memory and computational overhead. Through the synergistic integration of these strategies, EtinyNet-\acs{ssd} achieves a balance between efficiency and performance, allowing it to operate effectively within the resource constraints of \acp{mcu}. Notably, by employing EtinyNet as the backbone, EtinyNet-SSD achieves a remarkable \ac{map} of 56.4\% on the \ac{voc} dataset, underscoring the effectiveness of this model.

\noindent \textbf{XiNet-YOLOV7.} XiNet-YOLOV7, as introduced in~\cite{ancilotto2023}, presents a novel approach to optimizing \acp{nn} for \ac{od} on resource-constrained devices. The architecture builds upon the efficient operators identified through extensive analysis of standard, depthwise, and pointwise convolutions, ensuring maximum performance and energy efficiency. XiNet replaces depthwise convolutions with standard convolutions to enhance arithmetic intensity and reduce memory accesses, resulting in significant \ac{ram} and parameter savings. A critical innovation in XiNet is the use of \ac{has}~\cite{ancilotto2021}, which enables independent scaling of \ac{ram}, flash memory, and operations. This allows the \ac{dnn} architecture to be easily adapted to various hardware platforms, including \acp{mcu} and embedded devices. XiNet-YOLOV7 uses a hyper-parameter tuning approach involving $\alpha$, $\beta$, and $\gamma$, which scale the network's width, shape, and compression to balance between computational complexity, energy consumption, and performance. The use of standard convolutions, combined with efficient attention mechanisms and skip connections, results in a more hardware-efficient model without compromising on detection accuracy. The XiNet-YOLOV7 series "S", "M", and "L" are scaled variants optimized for different operational requirements, from low-energy consumption (XiNet-YOLOV7-S) to higher accuracy (XiNet-YOLOV7-L), each adapted for deployment on platforms with limited computational power. Using \ac{has}, the authors demonstrated the effectiveness of their method, as shown in Fig.~\ref{fig:mmacs_map}. These variants were deployed on an STM32H743 \ac{mcu} and achieved state-of-the-art performance, with XiNet-YOLOV7-"S", "M", and "L" achieving \ac{map} scores of 54\%, 67\%, and 74.9\% respectively, underscoring their suitability for real-time \ac{od} in \ac{tinyml} while significantly reducing energy and memory usage.
\begin{figure}[t]
{\includegraphics[width=0.6\textwidth]{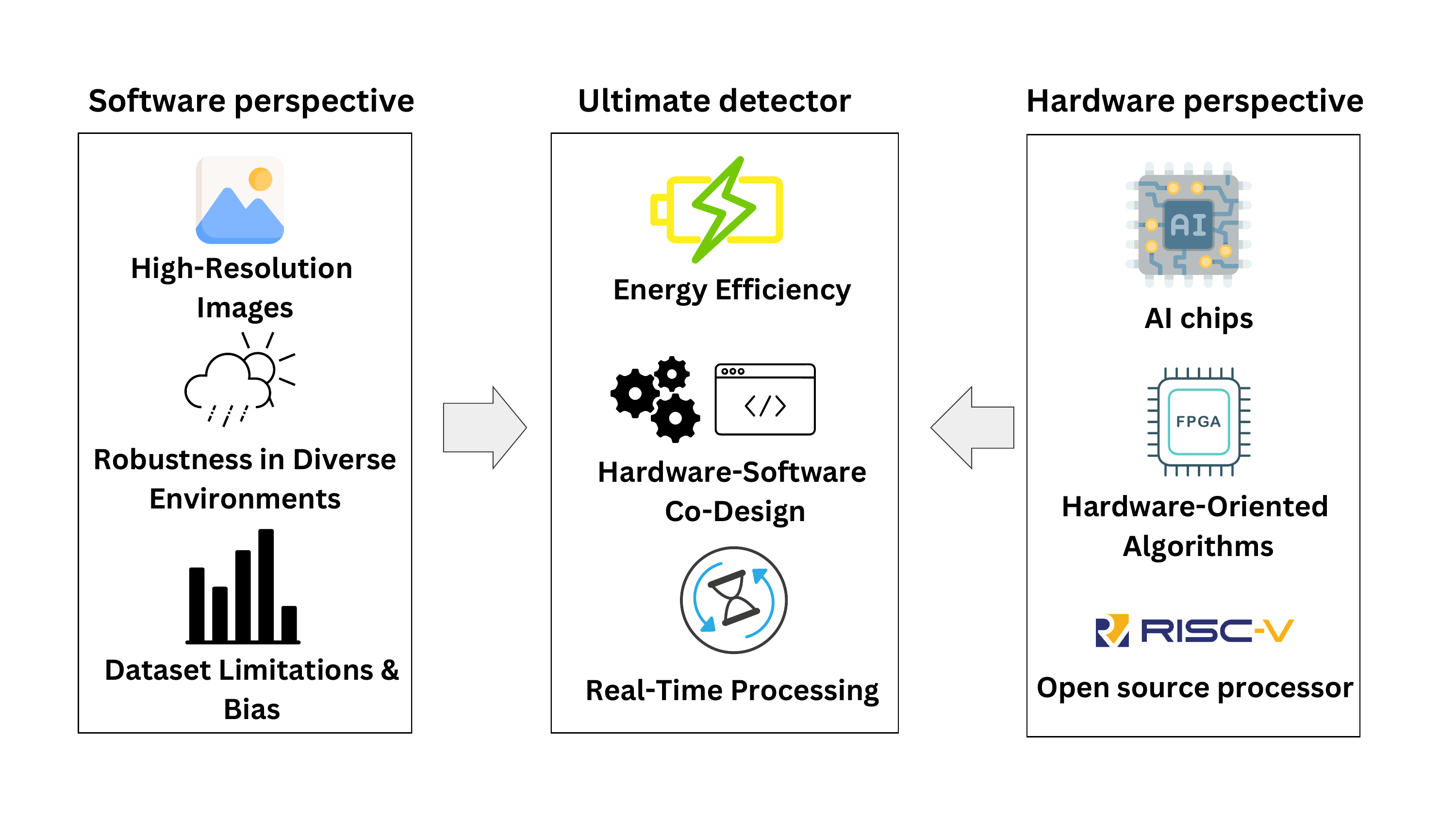}}
  \caption{Key challenges and considerations in developing the ultimate \ac{od} system, highlighting the software perspective (handling high-resolution images, robustness, and dataset limitations), hardware perspective (utilizing \ac{ai} chips, \acp{fpga}, and RISC-V architecture), and the need for energy-efficient, real-time hardware-software co-design.}
  \label{fig:challenges}
\end{figure}

\section{Open Challenges and Future Works}
\label{sec:object-detection-challenges}
\adcomment{In recent years, \ac{od} has seen significant advancements, with state-of-the-art models achieving impressive accuracy on large-scale datasets. However, deploying these models on embedded systems and \ac{tinyml} platforms remains an ongoing challenge due to stringent memory, computational, and energy constraints. The emergence of modern techniques, including transformer-based architectures, generative \ac{ai}, \acp{vlm}, and hardware-aware model optimizations, has opened new research avenues to address these limitations. Despite these innovations, ensuring real-time performance, efficiency, and adaptability of OD models in highly constrained environments still requires further exploration.}

In this section, we outline the open challenges associated with deploying \ac{od} on embedded systems and \ac{tinyml} platforms and propose potential research directions to address these challenges. These directions are categorized from two primary perspectives: software (managing high-resolution images, ensuring robustness, and overcoming dataset limitations) and hardware (advancements in AI chips, hardware-oriented algorithms, and open-source processors), as illustrated in Fig.~\ref{fig:challenges}. Additionally, the increasing role of \ac{nas}, \ac{kd}, and lightweight model compression techniques highlights the need for co-optimizing detection models with their hardware deployment. These techniques are particularly relevant given the shift towards multimodal and self-supervised approaches in \ac{od}, where models can leverage textual and contextual cues for more efficient and robust detection. We believe that the ideal detector will emerge from the synergy of both perspectives, addressing key challenges such as energy efficiency, real-time processing, and the integration of hardware-software co-design.

\noindent \textbf{1) Managing High-Resolution Images.} The challenge of balancing image resolution with computational feasibility remains a significant hurdle in embedded \ac{od} systems. Using lower-resolution images can lead to difficulties in detecting smaller or distant objects~\cite{chen2023tinydet, PowerofTiling}. Future research should explore adaptive resolution strategies that dynamically adjust image resolution based on the context and requirements of the \ac{od} task, allowing for efficient processing without compromising detection accuracy. \adcomment{Recent advances, such as \ac{esod}~\cite{liu2024esod}, demonstrate that small-object detection can be achieved by identifying and isolating high-interest regions at the feature level, rather than processing the full high-resolution image. This plug-and-play framework leverages adaptive slicing and sparse prediction heads to reduce computation and memory usage without degrading performance. Following this principle, future \ac{tinyml}-oriented methods could explore lightweight, context-aware patch-based processing to selectively activate only relevant spatial regions, making high-resolution processing feasible even on constrained \acp{mcu}.}

\noindent \textbf{2) Energy Efficiency.} Developing energy-efficient \ac{od} algorithms is essential for low-cost and battery-operated devices commonly used in \ac{iot} applications. Traditional \ac{dnn} models tend to be power-intensive, making them less suitable for resource-constrained environments. Promising innovations, such as Spiking \ac{yolo}, leverage \acp{snn} to significantly reduce power consumption while maintaining competitive performance~\cite{qu2023spikingneuralnetworkultralowlatency}. Future research should focus on optimizing \ac{snn}-based \ac{od} models and exploring hybrid architectures that integrate conventional and spiking neurons for enhanced energy efficiency. Implementing these models on neuromorphic hardware, such as the SpiNNaker chip~\cite{SpiNNaker}, further highlights the potential for real-time \ac{od} applications in resource-constrained environments due to their low power consumption and high efficiency.

\noindent \textbf{3) Robustness in Diverse Environments.} Ensuring reliable \ac{od} across various environmental conditions, such as changing lighting, weather, and occlusions, remains a major challenge~\cite{Li-obj}. Future work should investigate the development of adaptive \ac{od} models that can self-tune parameters in real-time, enhancing their robustness in diverse and dynamic environments. Techniques such as transfer learning and domain adaptation could also be explored to improve model performance across different scenarios. \adcomment{In addition, leveraging advanced data augmentation strategies such as photometric transformations, geometric distortions, and synthetic data generation can help models generalize better to unseen environments by simulating challenging real-world conditions during training~\cite{guo}.}

\noindent \textbf{4) Real-Time Processing.} Achieving low-latency \ac{od} is critical for applications such as autonomous navigation and real-time surveillance. The use of \acp{snn}, which mimic the human brain’s processing to achieve faster response times, shows potential~\cite{2019arXiv190306530K}. Future research could explore optimizing these networks for real-time OD tasks, as well as integrating hardware acceleration techniques to further reduce processing delays.

\noindent \textbf{5) Dataset Limitations and Bias.} The lack of extensive and diverse datasets specifically designed for embedded systems can result in biases in \ac{od} models, limiting their overall effectiveness~\cite{Syed}. Future research should focus on developing synthetic data generation techniques and augmentation methods that can enhance the diversity and quality of training datasets. Additionally, employing unsupervised and semi-supervised learning approaches could reduce the dependence on labeled data. Initiatives such as the Wake Vision dataset~\cite{Njor} represent a good start in this direction, offering a large-scale, high-quality dataset specifically tailored for \ac{tinyml} applications, setting a foundation for more robust and inclusive \ac{od} models.

\noindent \textbf{6) Hardware-Software Co-Design.} The integration of hardware and software plays a pivotal role in optimizing \ac{od} performance on embedded systems. System-level co-design strategies are essential for maximizing processing efficiency and minimizing power consumption~\cite{s23031185}. Leveraging \acp{llm} for architecture search, similar to the approach taken by GENIUS~\cite{genius}, can minimize the computational cost of searching for optimal architectures tailored for specific hardware platforms. By reducing the exploration space and discovering more efficient models quickly, \ac{llm}-driven approaches could significantly enhance the co-design process. \adcomment{In addition, hardware-software co-design frameworks, such as \ac{cfu} Playground~\cite{cfu}, offer an efficient way to tailor AI accelerators for \ac{tinyml} applications. \Ac{cfu} Playground enables rapid prototyping of FPGA-based \ac{ai} accelerators by allowing hardware designers to seamlessly integrate \acp{cfu} into the processor pipeline, facilitating the co-optimization of hardware and software, ensuring efficient execution of \ac{od} models on resource-constrained embedded platforms.}

\noindent \textbf{7) \Ac{ai} Accelerator Chips.} The adoption of specialized \ac{ai} accelerator chips, including \acp{npu} and \acp{asic}, \adcomment{ especially those integrated within modern \acp{mcu} such as the ARM Cortex-M55 STM32N6 with a built-in \ac{npu}, has significantly enhanced \acp{od} efficiency by providing dedicated hardware for high-speed inference in energy-constrained environments. Unlike general-purpose \acp{gpu} and \acp{tpu}, which are unsuitable for \ac{mcu}-class deployments, these embedded \ac{npu} solutions offer hardware-level acceleration tailored for low-latency, low-power inference.} The GnetDet model, optimized for a \acp{cnn} accelerator chip, has demonstrated high frames per second (FPS) on standard computing platforms~\cite{gnetdet-optimization}. As \ac{tinyml} applications demand increasingly efficient inference, future research should focus on optimizing \ac{od} models for various \ac{ai} accelerators while evaluating the impact of hardware configurations, memory hierarchies, and I/O interfaces on inference speed and accuracy. \adcomment{Beyond conventional \ac{ai} accelerator chips, emerging architectures such as neuromorphic and analog computing have gained traction for energy-efficient machine learning in resource-constrained environments~\cite{qu2023spikingneuralnetworkultralowlatency}. Neuromorphic chips, inspired by biological neural systems, enable event-driven processing through \acp{snn}, significantly reducing power consumption compared to traditional \ac{dnn} accelerators. Similarly, analog computing leverages in-memory computation~\cite{Chuteng} to perform matrix operations with minimal energy overhead, making it particularly suitable for ultra-low-power \ac{tinyml} applications. Integrating these architectures into \ac{od} pipelines could enable real-time, low-latency inference on edge devices with stringent power and memory constraints. }

\noindent \textbf{8) Hardware-Aware Co-Design and RISC-V Opportunities.} Achieving real-time, ultra-efficient \ac{od} on embedded platforms requires tight coupling between algorithm design and hardware capabilities. Rather than adapting general-purpose algorithms to constrained devices, recent trends emphasize hardware-oriented models developed with specific architectural constraints in mind~\cite{FENG2019309}. In particular, multi-objective \ac{hnas} techniques are gaining traction by jointly optimizing neural architectures for both computational efficiency and hardware cost~\cite{HNAS}. This co-design philosophy is further enabled by the rise of open-source instruction set architectures such as RISC-V~\cite{riscv}, which offer modularity, flexibility, and low-cost customization for embedded \ac{ai}. Recent efforts like Flex-RV~\cite{Ozer2024} showcase how lightweight RISC-V-based microprocessors can integrate machine learning accelerators directly into their pipeline, paving the way for task-specific \ac{tinyml} solutions. These developments suggest a promising direction where open hardware ecosystems and co-designed algorithms together drive scalable, energy-efficient \ac{od} at the edge.

\noindent \adcomment{\textbf{9) Designing Transformer-Based \ac{od} for \ac{tinyml}.}
Transformer-based models have achieved state-of-the-art performance in various computer vision tasks, surpassing traditional \acp{cnn}\cite{vit}. This trend has extended to \ac{od}, with the recent YOLOv12 architecture incorporating attention mechanisms to enhance feature extraction. YOLOv12's Nano variant achieves superior results with fewer parameters, showcasing the potential of attention-based models for resource-constrained environments~\cite{tian2025yolov12}. However, deploying transformer-based object detectors on \ac{mcu}-class devices presents challenges due to their computational complexity and memory requirements. Unlike \acp{cnn}, where computations are localized, transformers operate with self-attention mechanisms that scale quadratically with input size, making them inefficient for platforms with limited \ac{sram} and energy budgets. Emerging lightweight transformer architectures, such as MobileViT~\cite{mehta2021mobilevit}, TinyViT~\cite{wu2022tinyvit}, and EdgeFormer~\cite{ge2022edgeformer}, have shown improvements for edge devices, but their applicability to \ac{tinyml} remains largely unexplored, as their model footprints and computational demands often exceed the capabilities of typical \acp{mcu}. To address these limitations, future work should focus on optimizing transformers for \ac{tinyml}. One promising direction involves \ac{kd} to transfer the representations learned by large-scale transformer detectors into compact models tailored for \ac{mcu} deployment. Through \ac{kd}, a compact student model can retain the expressiveness of transformers while operating efficiently under stringent memory constraints. In addition, recent work such as MCUFormer~\cite{liang2023mcuformer} demonstrates the feasibility of transformer-based models on embedded platforms by employing techniques such as low-rank decomposition, token overwriting, and quantized attention mechanisms. MCUFormer achieves 73.6\% accuracy on ImageNet while running on a Cortex-M7 \ac{mcu} with only 320KB of available memory, underscoring the potential for transformer-based \ac{od} on \ac{tinyml}. Further research should adapt such designs for \ac{od} pipelines, ensuring alignment between model architectures and hardware capabilities. Exploring hybrid transformer-\ac{cnn} models and applying transformers to remote sensing tasks may offer additional insights into optimizing these models for \ac{tinyml}. To enable efficient deployment of transformer-based \ac{od} models, new approaches to self-attention, such as Linformer~\cite{wang2020linformer} and Performer~\cite{choromanski2020rethinking}, should be explored to reduce memory overhead while maintaining global feature modeling. As an alternative approach, \ac{nas} presents an opportunity to identify optimal transformer configurations tailored to resource constraints, dynamically adjusting architectural components to fit within \ac{mcu} limitations. Furthermore, specialized operator libraries like CMSIS-NN and TensorFlow Lite Micro could accelerate inference while maintaining power efficiency. The intersection of transformer methods, \ac{kd}, hardware-aware model search, and efficient inference scheduling is a critical research frontier for achieving high-performance, energy-efficient \ac{od} on embedded platforms.}

\noindent \adcomment{\textbf{10) Expanding Capabilities: Multimodal Learning and Generative \ac{ai} in \ac{od}.}
In addition to architectural streamlining for transformers, the next frontier in \ac{tinyml} \ac{od} lies in enhancing object detectors with richer representational capabilities through generative and multimodal \ac{ai}. Generative \ac{ai} techniques, such as diffusion models~\cite{zhu2024odgen} and \acp{gan}~\cite{jung2024study}, offer powerful tools to improve training data quality, mitigating data scarcity and enhancing the generalization of small-scale detectors. These approaches are especially beneficial for low-shot learning, where limited annotated samples are available for embedded systems. Furthermore, multimodal \ac{kd} presents a promising strategy for transferring representations from transformer-based models to more compact \ac{tinyml}-friendly detectors. By integrating cross-domain knowledge from \acp{vlm}~\cite{zhang2024vision} and large-scale transformers\cite{vit}, multimodal \ac{kd} allows \ac{tinyml} models to retain the contextual reasoning of more complex architectures while remaining computationally efficient. This approach extends beyond standard distillation, which focuses mainly on reducing model size, and incorporates a broader knowledge transfer that improves the model's expressive power. Another important direction involves developing compression techniques specifically for transformer-based \ac{od} models in \ac{tinyml}. While traditional compression techniques have shown success with edge devices, they must be adapted to \ac{tinyml} environments, where even "lightweight" models can exceed the available \ac{sram} and computation budgets. \ac{nas} could help identify optimized transformer architectures that balance model expressiveness and hardware feasibility. Emerging zero-shot \ac{vlm} models, such as SpatialLM~\cite{spatiallm}, provide capabilities to extend \ac{tinyml} \ac{od} to open-set scenarios, allowing models to detect novel object categories without retraining. This could be a game-changer for \ac{tinyml} \ac{od}, enabling more flexible and scalable detection systems in dynamic environments. To truly capitalize on these advancements, future research must focus on integrating generative \ac{ai}, multimodal learning, and transformer compression into the \ac{tinyml} domain. This includes rethinking transformer compression, developing hardware-aware architecture design, and enabling efficient cross-domain knowledge transfer to achieve real-time, energy-efficient \ac{od} on ultra-low-power embedded platforms.}

\section{Conclusion}
\label{sec:conclusion}
In this paper, a comprehensive survey of model compression techniques such as quantization, pruning, \ac{kd}, and \ac{nas}, tailored to enhance \ac{od} performance in embedded systems, was conducted. These techniques were analyzed in the context of their application to consumer electronics, \ac{iot}, and edge computing, highlighting their effectiveness in optimizing model efficiency while maintaining accuracy. Key metrics for assessing optimization success were also reviewed, and comparative insights into the practical impacts of these methods were provided. Additionally, existing challenges in deploying \ac{od} models in resource-constrained environments were outlined, with an emphasis on the need for further advancements in energy efficiency and real-time processing. The findings of this survey suggest promising avenues for future research, particularly in developing more energy-efficient algorithms and exploring integrated hardware-software solutions to enhance \ac{od} capabilities in embedded systems.

\bibliographystyle{ACM-Reference-Format}
\bibliography{acmart}
\end{document}